\documentclass[final]{article}

\PassOptionsToPackage{numbers, compress}{natbib}

\usepackage{neurips_2025}

\usepackage[utf8]{inputenc} 
\usepackage[T1]{fontenc}    
\usepackage{url}            
\usepackage{amsfonts}       
\usepackage{nicefrac}       
\usepackage{microtype}      
\usepackage{xcolor}         
\usepackage{booktabs}       
\usepackage{colortbl}
\usepackage{amsmath}
\usepackage{multirow}
\usepackage{graphicx}   
\usepackage{array} 
\usepackage{amsmath}
\usepackage{adjustbox}
\usepackage{makecell} 
\usepackage{enumitem}

\usepackage{tabularx}
\usepackage{caption}  
\usepackage{subcaption}
\usepackage{graphicx}
\usepackage{amsmath}
\usepackage{amssymb}
\usepackage{booktabs}
\usepackage{CJKutf8}
\usepackage{overpic}
\usepackage{wrapfig}
\usepackage{enumitem}

\definecolor{mycolor_blue}{HTML}{E7EFFA}
\definecolor{mycolor_green}{HTML}{E6F8E0}
\definecolor{mycolor_gray}{HTML}{ECECEC}
\definecolor{pearDark}{HTML}{2980B9}
\definecolor{citecolor}{HTML}{2980b9}
\definecolor{linkcolor}{HTML}{c0392b}
\definecolor{sem}{HTML}{2E75B6}
\definecolor{tok}{HTML}{F3B000}
\usepackage[hidelinks,breaklinks=true,colorlinks,citecolor=citecolor,linkcolor=linkcolor]
{hyperref} 

\makeatletter
  \newcommand\figcaption{\def\@captype{figure}\caption}
  \newcommand\tabcaption{\def\@captype{table}\caption}
\makeatother
\usepackage{amsmath}
\usepackage{amssymb}
\usepackage{mathtools}
\usepackage{amsthm}
\usepackage{makecell}
\usepackage{multirow}
\usepackage{ulem}
\usepackage{xspace}

\newcommand{\tablefontsizeone}{\fontsize{10}{13}\selectfont} 
\newcommand{\tablefontsizetwo}{\fontsize{8}{10}\selectfont} 

\newcommand{\tablefontsizethree}{\fontsize{22}{28}\selectfont} 

\title{EvoMoE: Expert Evolution in Mixture of Experts for Multimodal Large Language Models }

\begin{document}

\author{Linglin Jing\textsuperscript{\rm 1}\thanks{This work was completed during his internship at Ant Group.}, Yuting Gao$^{2}$, Zhigang Wang$^{1}$, Wang Lan$^{2}$, \\ Yiwen Tang$^{1}$, Wenhai Wang$^{1}$, Kaipeng Zhang$^{1}$,Qingpei Guo\textsuperscript{\rm 2}\thanks{Corresponding author} \\
$^{1}$Shanghai AI Laboratory,\\
$^{2}$Ant Group
\\
{\tt\small \{l.jing@lboro.ac.uk}\}
}

\maketitle

\begin{abstract}
Recent advancements have shown that the Mixture of Experts (MoE) approach significantly enhances the capacity of large language models (LLMs) and improves performance on downstream tasks. Building on these promising results, multi-modal large language models (MLLMs) have increasingly adopted MoE techniques. However, existing multi-modal MoE tuning methods typically face two key challenges: \textbf{\textit{expert uniformity}} and 
\textbf{\textit{router rigidity}}. Expert uniformity occurs because MoE experts are often initialized by simply replicating the FFN parameters from LLMs, leading to homogenized experts and weakening the intended diversification of the MoE architecture. Meanwhile, router rigidity stems from the prevalent use of static linear routers for expert selection, which fail to distinguish between visual and textual tokens, resulting in similar expert distributions for image and text. To address these limitations, we propose EvoMoE, an innovative MoE tuning framework. EvoMoE introduces a meticulously designed expert initialization strategy that progressively evolves multiple robust experts from a single trainable expert, a process termed expert evolution that specifically targets severe expert homogenization. Furthermore, we introduce the Dynamic Token-aware Router (DTR), a novel routing mechanism that allocates input tokens to appropriate experts based on their modality and intrinsic token values. This dynamic routing is facilitated by hypernetworks, which dynamically generate routing weights tailored for each individual token. Extensive experiments demonstrate that EvoMoE significantly outperforms other sparse MLLMs across a variety of multi-modal benchmarks, including MME, MMBench, TextVQA, and POPE. Our results highlight the effectiveness of EvoMoE in enhancing the performance of MLLMs by addressing the critical issues of expert uniformity and router rigidity.

\end{abstract}

\section{Introduction}
\label{sec:Intro}

Multi-modal Large Language Models (MLLMs), such as GPT-4~\cite{achiam2023gpt} and Llama 3~\cite{grattafiori2024llama}, have achieved significant success in addressing open-world tasks, thanks to their scaled-up architectures. However, scaling up models often increases computational demands and is limited by device capacity. To address these challenges, sparsely activated mixture-of-expert (MoE) models have gained popularity in large language models (LLMs)~\cite{zhao2024hypermoe,liu2024survey,zhai2023smartmoe,rajbhandari2022deepspeed,zhu2024llama,zhong2024adapmoe}, reducing computational costs and enhancing efficiency. MoE models typically include multiple experts and a routing network that selects the optimal expert for each input token. This design minimizes interference among diverse input tokens, enabling each expert to specialize more effectively in specific tasks. For example, DeepSeek V3~\cite{liu2024deepseek} employ MoE language models with 671B parameters, activating 37B parameters, and have achieved notable results.

\begin{figure}[h!]
    \centering
    \begin{subfigure}{0.490\textwidth}
        \centering
        \includegraphics[width=\linewidth]{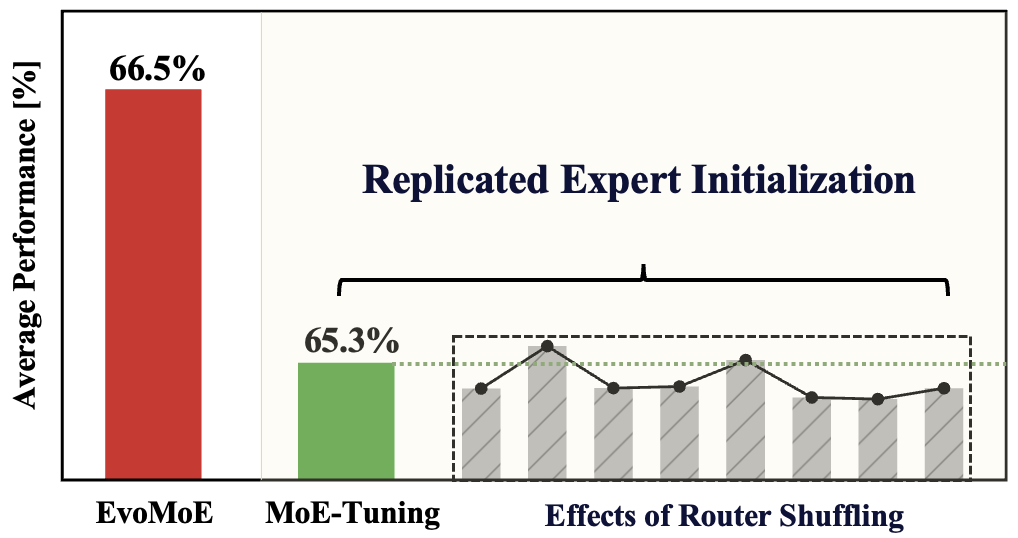} 
        \caption{Expert Uniformity} 
        \label{fig:expertU}
    \end{subfigure}
    \hfill
    \begin{subfigure}{0.49\textwidth}
        \centering
        \includegraphics[width=\linewidth]{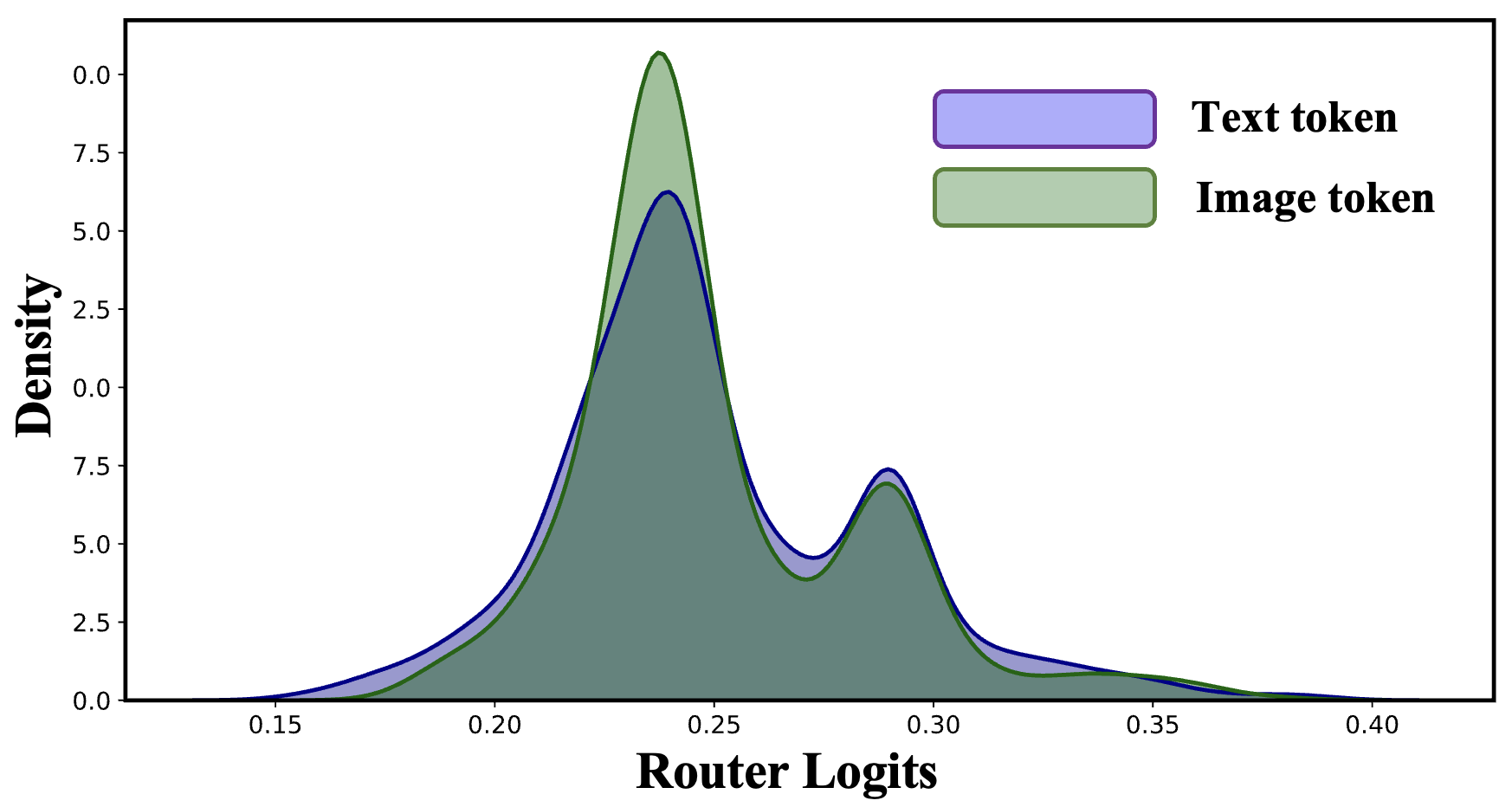} 
        \caption{Router Rigidity} 
        \label{fig:RouterL}
    \end{subfigure}
   \caption{Two key challenges in MoE-tuning. (a) \textbf{Expert Uniformity}: Randomly shuffling the router during inference results in negligible performance degradation, suggesting uniformity among experts derived from replicated initialization. (b) \textbf{Router Rigidity}: Kernel density estimation (KDE) of the logits for image and text tokens reveals that the linear router generates input-insensitive selections, leading to static distributions with significant overlap in the density of logits for image and text token.}
    \label{fig:two_images}
\end{figure}

The success of MoE in LLMs has spurred significant interest in its application to MLLMs~\cite{lee2024moai,lin2024moe,liu2024deepseek,rang2025eve,yang2024qwen2technicalreport,zhang2024hyperllava, zong2024mova}. A recent prominent approach is MoE-LLaVA~\cite{lin2024moe}, which introduces MoE-tuning, a multi-stage process that converts dense MLLM models into sparse MoE structures during instruction tuning, specifically tailored for multi-modal tasks.
However, in MoE-tuning, experts are typically initialized by replication, leading to the first critical challenge: \textit{\textbf{expert uniformity}}. This results in expert homogenization during multi-stage tuning, thereby impeding specialization and undermining the efficacy of the MoE framework. Figure~\ref{fig:expertU} illustrates an experiment on expert uniformity, in which we repeatedly shuffled the logits across router layers during evaluation and found no significant drop in average performance. This indicates that experts, which were initially replicated from a common source, tend to become homogeneous after training rather than developing specialized functions. This finding contradicts the fundamental principle of the Mixture-of-Experts (MoE) approach, which aims to enhance task-specific performance through the use of diverse experts.
Additionally, MoE-tuning commonly employs a simple linear layer for expert assignment, mimicking the approach used in LLMs. This leads to the second challenge: \textit{\textbf{router rigidity}}. The shared linear router struggles to differentiate between visual and text tokens in MLLMs, resulting in uniform predictions. This limits the model's adaptability and effectiveness in multi-modal tasks. Figure~\ref{fig:RouterL} illustrates this rigidity using Kernel Density Estimation (KDE) plot, revealing significant overlap in the logit distributions of image and text tokens. This overlap indicates that the router becomes inflexible during training, producing uniform output distributions regardless of the input type, thus restricting the model's adaptability in multi-modal tasks.

To tackle the aforementioned challenges, we introduce \textbf{EvoMoE}, a sparse MoE framework tailored for MLLM.  
Our method builds on MoE-tuning framework, gradually transforming dense models into MoE structures. To address the challenge of \textit{\textbf{expert uniformity}} and enhance the diversity of expert initialization, we introduce a novel approach called Expert Evolution, which generates diverse MoE experts by iteratively adapting expert parameters through a dynamic evolution value. This evolution value integrates prior expertise with gradient-based updates, enabling continuous refinement and evolution. Consequently, the technique evolves multiple robust MoE experts from a single trainable expert.
%
To address \textit{\textbf{router rigidity}} and enhance the connection between the router and input modalities, we propose the Dynamic Token-aware Router (DTR). This router dynamically allocates input tokens to specific experts. Specifically, we employ a hypernetwork to generate unique parameters for each router, tailored to the unique value of each token.

Our core contributions are summarized as follows:

\begin{itemize}[left=0pt]

\item We introduce EvoMoE, an innovative MoE-tuning framework specifically designed for MLLMs, which effectively addresses two critical challenges: expert uniformity and router rigidity.

\item To tackle expert uniformity, we propose a novel method termed expert evolution, which flexibly generates a diverse set of MoE experts. Furthermore, to mitigate router rigidity, we introduce DTR, which assigns input tokens to specific experts based on their modality and intrinsic value.

\item Extensive experiments on language models of various sizes demonstrate that EvoMoE achieves better performance with fewer activated parameters.
\end{itemize}

\section{Related Works}
\label{sec:Related}

\subsection{multi-modal Large Language Model.}

LLMs~\cite{liu2024deepseek,yang2024qwen2technicalreport,thirunavukarasu2023large,naveed2023comprehensive,chang2024survey} have demonstrated outstanding capabilities in reasoning, comprehension, and question answering. Building on this, recent studies have extended LLMs into the visual domain, leading to the creation of MLLMs.
LLaVA 1.5~\cite{liu2023improvedllava} marks a notable advancement in MLLMs by integrating visual and textual modalities using a simple yet effective architecture, employing a pre-trained vision encoder and language model linked by a lightweight projection layer, achieving strong performance across multi-modal tasks. 
Recent MLLMs continue to push the boundaries of vision-language integration through novel architectures. For instance, InternVL~\cite{chen2024internvl} enhances fine-grained visual-semantic alignment by decomposing high-resolution images into regional patches with a dynamic multi-scale module and fusing features through pixel-shuffle-based method.
Qwen2.5-VL~\cite{bai2025qwen2} combines the Qwen language model~\cite{bai2023qwen} with refined vision-language alignment techniques, using fine-grained multi-modal attention and dynamic adaptation to excel in various multi-modal benchmarks.
%
Meanwhile, scaling efforts in systems like GPT-4o~\cite{hurst2024gpt}, Gemini~\cite{team2024gemini}, and DeepSeek-VL~\cite{lu2024deepseek} highlights the importance of model and data scaling, with expanded architectures showing emergent capabilities in multi-modal reasoning, code synthesis, and long-context comprehension.

\subsection{Mixture of Experts in Multi-modal Learning.}

Given the substantial computational overhead associated with training and deploying MLLMs, researchers are increasingly turning to the MoE architecture to enhance efficiency. This approach can be broadly categorized into two main strategies: one directly integrates multi-modal capabilities into an LLM with an MoE architecture, as seen in MLLMs like Qwen2.5-VL~\cite{bai2025qwen2}, Kimi-VL~\cite{team2025kimi} and MiniMax-VL~\cite{li2025minimax}, leveraging its inherent efficiency and scalability. The other strategy extends a dense LLM to an MoE-based architecture, offering greater flexibility in adapting to diverse tasks and modalities while maintaining computational efficiency.
For instance, HyperLLaVA~\cite{zhang2024hyperllava} integrates additional visual experts within the vision encoder and the LLM. These hypernet-based experts dynamically capture input characteristics, thereby offering improved feature extraction capabilities within these components.
LLaVA-MoLE~\cite{chen2024llava} creates a set of LoRA experts and a linear router for the FFN layer to mitigate data conflicts when combining multiple distinct instruction datasets.
Recently, MoE-LLaVA~\cite{lin2024moe} present MoE-tuning, a novel three-stage framework that progressively converts the dense FFN layers in LLMs into a MoE structure. This method incorporates a linear router, significantly lowering the required activation parameters while maintaining or even surpassing the performance of dense models.
Building on MoE-tuning, Eve~\cite{chen2024eve} incorporates an extra visual expert in stage III of the LLM to distinguish image tokens from text tokens. 
However, these MoE-tuning approaches encounter challenges concerning Expert Uniformity and Router Rigidity.

\section{Methods}
\label{sec:MD}

In this paper, we present EvoMoE, a novel MoE-tuning approach addressing expert uniformity and router rigidity.
Section~\ref{subsec:Framework} overviews the three-stage tuning framework of EvoMoE.
In Section~\ref{subsec:Evo}, we introduce the expert evolution strategy for generating diverse MoE experts during Stage II.
Finally, Section~\ref{subsec:DMR} details the Stage III routing mechanism, which dynamically assigns input tokens to suitable experts based on modality and inherent token values.




\begin{figure}[!t] 
\begin{center}
\includegraphics[width=1.0\linewidth]{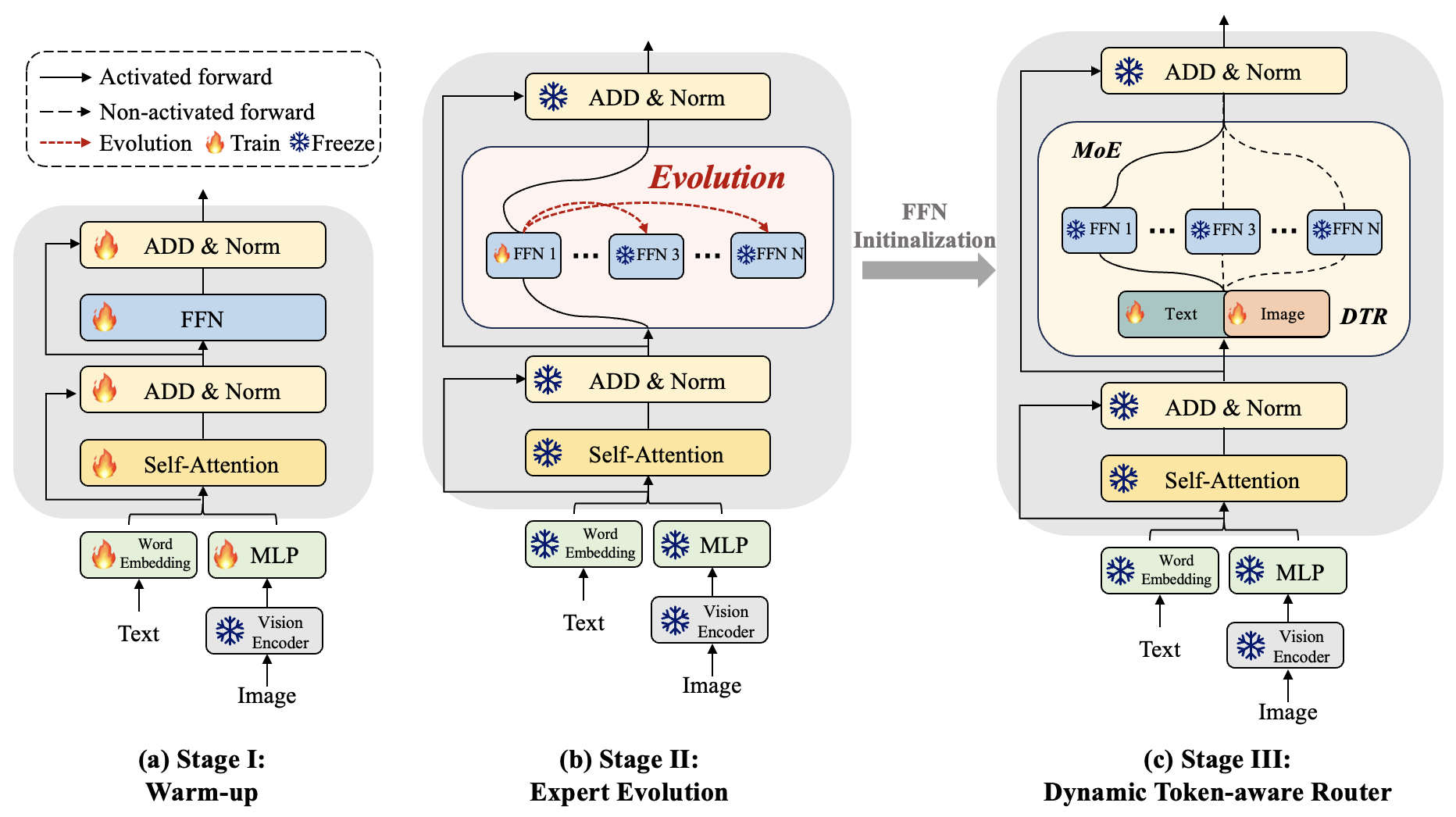}
\end{center}
\caption{\textbf{The framework of EvoMoE.} EvoMoE comprises three stages of instruction-tuning: (a) Warm-up: Begin training with multi-modal instruction data to familiarize the model with understanding capabilities, utilizing parameters initialized during the MoE-LLaVA~\cite{lin2024moe} pretraining stage. (b) Expert Evolution: Train only FFN1, while evolving other experts from FFN1, and (c) Dynamic Token-aware Router: Use FFNs evolved in Stage II for expert initialization and incorporate the DTR for MoE, with only the DTR trainable during this stage, while all other parameters remain frozen.}
\label{fig:overview}
\end{figure}

\subsection{Framework Overview}
\label{subsec:Framework}
In the MoE-tuning approach, the initial pre-training phase establishes cross-modal alignment by utilizing an MLP projector to map image tokens into the LLM's latent space, equipping the model with foundational visual-language understanding capabilities. Leveraging this groundwork, we introduce EvoMoE, building upon the pre-training phase of MoE-LLaVAA~\cite{lin2024moe} as our initialization base. Our methodology further advances this foundation through a three-stage framework designed to systematically evolve the dense pre-trained backbone into a sparse MoE architecture.
This framework incorporates two key components: (1) Expert Evolution: Gradually generate new experts using an evolution strategy. (2) Dynamic Token-aware Routing: Implement a dynamic routing mechanism that prioritizes input-relevant experts.
%
%
Figure~\ref{fig:overview}, coupled with the subsequent description, clarifies the details of the three-stage framework:

\textbf{\textit{Stage I: Understanding Warm-up.}} 
To equip the model with basic instruction-following capabilities, we utilize a collection of instruction datasets to train all parameters of the dense LLM and the corresponding Multi-Layer Perceptron (MLP) layer. 

\textbf{\textit{Stage II: Expert Evolution.}} In this stage, we introduce a novel methodology for constructing Mixture-of-Experts (MoE) experts, where each expert is instantiated as a unique Feed-Forward Network (FFN) layer within the LLM. By employing expert evolution, a process that dynamically balances prior expertise with gradient-based updates, we progressively derive multiple robust FFN experts from a single trainable FFN during training. This iterative approach enables continuous refinement and adaptation, enhancing the diversity, specialization, and robustness of the experts. As a result, the model’s adaptability and overall performance are significantly improved.

%

%

\textbf{\textit{Stage III: Dynamic Token-aware Router.}} In this stage, we introduce the Dynamic Token-aware Router (DTR), which dynamically allocates input tokens to appropriate experts based on their modality. The DTR parameters are generated by a hypernetwork, creating routing decisions specifically tailored for each input token. The experts are initialized through the evolutionary process described in Stage II.

\subsection{MoE Expert Evolution}
\label{subsec:Evo}

The existing MoE-tuning approach typically replicates the FFN parameters to initialize multiple MoE experts, resulting in expert uniformity issues during training.
To address this challenge, as illustrated in Figure~\ref{fig:overview} (b), we propose a novel initialization strategy, expert evolution, which gradually evolves new experts by dynamically balancing prior expertise with gradient-based updates:
\begin{align}
\theta_n \leftarrow \beta \cdot \theta_1+(1-\beta) \cdot {\nabla \theta_1},
\end{align}
where $\theta_1$ represents the network parameters of the original trainable FFN, initialized using the output of stage I and designated as expert 1.
$\theta_n$ denotes the newly generated FFN experts, where 
the index $n=[2,3,\cdots,N]$ represents each of the $N$ experts. Meanwhile, $\nabla \theta_1$ represents the gradient update for the trainable expert. 
$\beta$ denotes the evolution value within the range [0,1] which controls the evolution rate. A larger $\beta$ emphasizes historical data, producing a smoother average by diminishing the impact of recent changes. In our experiments, $\beta$ is randomly assigned a value within a specified range at each training step for improved generalization, as detailed in Section~\ref{sec:setup}.

\begin{wrapfigure}{r}{0.45\textwidth}
  \vspace{-18pt}
  \begin{center}
    \includegraphics[width=0.45\textwidth]{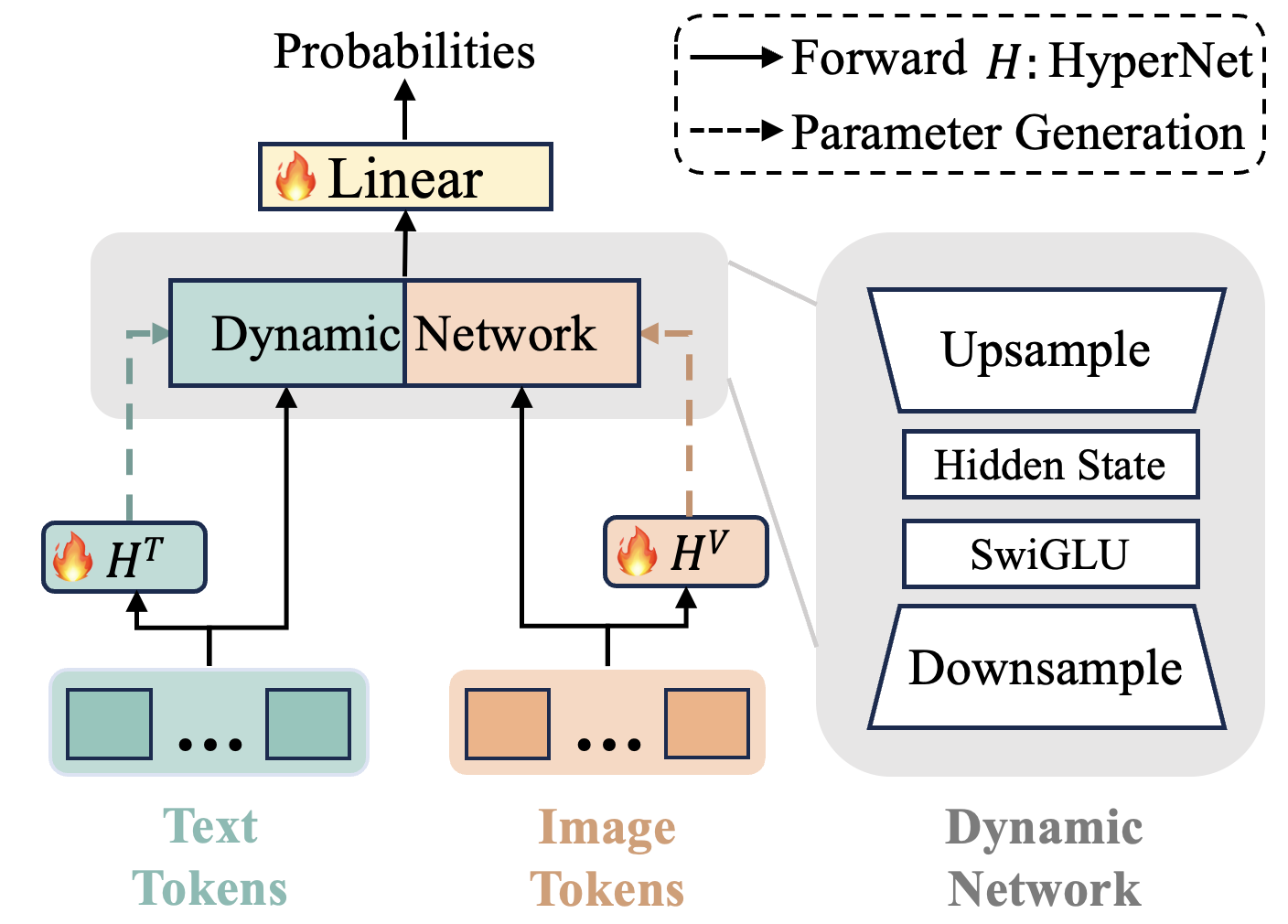}
  \end{center}
  \caption{\textbf{Dynamic Token-aware Router (DTR).} Two input-guided hypernetworks dynamically generate network parameters for the up-sampling and down-sampling layers based on visual and text tokens. The final linear layer predicts probabilities and selects the top-k experts. In this module, only the hypernetworks and the linear layer are trainable.}
  \vspace{-24pt}
  \label{figure:DTR_Evo}
\end{wrapfigure}
We exclusively train $\theta_1$ of expert 1, allowing it to evolve into different experts through varying evolution rates $\beta$. Importantly, these evolved experts and all other LLM and MLP parameters remain frozen during training.

\subsection{Dynamic Token-aware Router (DTR)}
\label{subsec:DMR}

In the MoE-tuning approach, a shared linear router selects the top K experts for both visual tokens  ${V}=\left[v_1, v_2, \cdots, v_P\right] \in \mathbb{R}^{P \times C}$ and text tokens ${T}=\left[t_1, t_2, \cdots, t_M\right] \in \mathbb{R}^{M \times C}$, here, $P$ is the sequence length of visual tokens, $M$ is the sequence length of text tokens, and $C$ denotes the hidden size of the LLM. 
This setup might not differentiate between visual and text tokens, limiting effectiveness in multi-modal tasks and resulting in uniform predictions, known as router rigidity.

%

To address this challenge, we introduce the Dynamic Token-aware Router (DTR) as a key component of the EvoMoE framework. EvoMoE consists of $L$ blocks, each integrating multi-head self-attention (MSA), feed-forward neural networks (FFN), layer normalization (LN) and residual connections. The framework is structured as follows:
\begin{gather} 
\label{eqn:sw}
{z}_0^V=\left[v_1, v_2, \cdots, v_P\right], \quad {z}_0^T=\left[t_1, t_2, \cdots, t_M\right], \quad l=1 \ldots L.  \\
{z}_{l}^{V\prime}=\operatorname{MSA}\left(\operatorname{LN}\left({z}_{l-1}^V\right)\right)+{z}_{l-1}^V, \quad {z}_{l}^{T\prime}=\operatorname{MSA}\left(\operatorname{LN}\left({z}_{l-1}^T\right)\right)+{z}_{l-1}^T. \\
{z}_{l}^V=\operatorname{FFN}(\operatorname{DTR}\left(\operatorname{LN}\left({z}_{l}^{V\prime}\right)\right))+{z}_{l}^{V\prime}, \quad {z}_{l}^T=\operatorname{FFN}(\operatorname{DTR}\left(\operatorname{LN}\left({z}_{l}^{T\prime}\right)\right))+{z}_{l}^{T\prime}.
\end{gather}
%
The architecture of DTR, as depicted in Figure~\ref{figure:DTR_Evo}, innovatively manages input visual and text tokens through two hypernetworks, denoted as $\mathcal{H}^V$ and $\mathcal{H}^T$. Each hypernetwork consists of two MLP layers, allowing it to generate adaptive parameters customized to each token. For instance, when processing visual tokens $V$, the hypernetwork $\mathcal{H}^V$ predicts modality-specific parameters $\Theta^{V}$, optimized for visual inputs, as detailed below:
\begin{align} \label{eqn:hyper}
& \Theta^{V}=\left(w_1 \boldsymbol{z}^{\prime(V)}+b_1\right) w_2+b_2,
\end{align}
where $w1$ and $w2$ denote the weights for two MLPs in $H^V$, while $b_1$ and $b_2$ represent the corresponding biases. Similarly, $\Theta^{T}$ denotes the dynamic weights for the text token input, which are generated in a manner analogous to that of the visual.

DTR consists of a pair of down-sampling and up-sampling layers designed to dynamically extract visual and textual information from various input tokens.
Finally, the prediction of expert probabilities $\rho$ is formulated as follows:
%
\begin{align}
& \Theta_{\text {up}}^\tau, \Theta_{\text {down }}^\tau=\mathcal{H}^\tau\left(z^{\tau\prime}\right), \text { where } \tau \in V, T \\\hspace{2em}
& \mathcal{E}^\tau=\Theta_{\text {up }}^\tau\left(\operatorname{SwiGLU}\left(\Theta_{\text {down }}^\tau\left(z^{\tau\prime}\right)\right)\right), \\
&  \rho^\tau=\left(\phi\left(\mathcal{E}^\tau\right)\right),
\end{align}
where $\tau$ denotes the visual and text input tokens. $z^{\prime}$ is the output of a single MSA block.
$\Theta_{up}$ and $\Theta_{down}$ correspond to the weights of the up-sampling and down-sampling layers, respectively, which are generated by the hypernetwork  $\mathcal{H}$. In this context, the SwiGLU activation function is utilized. The symbol $\phi$ denotes an MLP layer that serves as the final router. During training, only $\mathcal{H}^V$, $\mathcal{H}^T$, and $\phi$ are fine-tuned. Each token is processed by the top K experts with the highest probability. 

\subsection{Training Objective}

In alignment with~\cite{lin2024moe}, the overall loss function of EvoMoE consists of two components: the regression loss:  $\mathcal{L}_{\text {regressive }}$ and the auxiliary loss $\mathcal{L}_{\text {aux}}$. Regression loss is designed to optimize model performance, while auxiliary loss aims to promote a balanced load distribution across the router $\phi$:
\begin{equation}
\mathcal{L}_{\text {total }}=\mathcal{L}_{\text {regressive }}+\alpha \cdot \mathcal{L}_{\text {aux }.}
\end{equation}
Here, $\alpha$ is a hyperparameter that controls the weight of the auxiliary loss and is set to 0.001 during the training process. The detailed formulas are provided in the supplementary materials.

\section{Experiments}
\label{sec:exp}

\subsection{Experiments Setup}
\label{sec:setup}

\textbf{Model Details.}
EvoMoE is built on the MoE-tuning and LLaVA 1.5 frameworks, centering on the Evolution Strategy and the Dynamic Token-aware Router (DTR). The training framework suits various sizes, with experiments on LLMs with 0.5B, 1.8B, 2.7B, and 7B parameters showing strong generalization.
Importantly, EvoMoE achieves state-of-the-art performance by activating only the top-1 expert, which offers a significant advantage in terms of the number of activated parameters.

\textbf{Training Datasets.}
%
In Stage I, following MoE-LLaVA~\cite{lin2024moe}, we use a diverse dataset collection, including MIMIC-IT~\cite{li2023mimic}, LRV~\cite{liu2023aligning}, SViT~\cite{zhao2023svit}, and LVIS~\cite{wang2023see}, to enhance the MLLM’s general multi-modal comprehension skills.
Stage II employs the LLaVA-mix-665k~\cite{lai2024lisa} dataset to develop evolution experts.
In Stage III, the same LLaVA-mix-665k dataset is used to train the DTR.

\textbf{Evaluation.}
We evaluate the effectiveness and robustness of EvoMoE across diverse scenarios through performance evaluations on an extensive range of multi-modal benchmarks, including VQA-v2~\cite{goyal2017making}, GQA~\cite{hudson2019gqa}, SQA~\cite{lu2022learn}, TextVQA~\cite{singh2019towards}, POPE~\cite{li2023evaluating}, MME~\cite{Fu2023MMEAC}, and MMBench~\cite{liu2024mmbench}.

\textbf{Implementation Details.}
In our experiments, CLIP-L~\cite{radford2021learning} and SigLIP-L~\cite{zhai2023sigmoid} were utilized as the image encoders. Throughout all experiments, the batch size was consistently maintained at 4, with a gradient accumulation of 2. For all the three stages of instruction tuning, the initial learning rate was 2e-5, and we consistently select the top-1 expert across all experiments.
In the evolution strategy, the evolution rate is randomly chosen from one of three specified ranges at each training: [0.9–0.99], [0.8–0.89], and [0.7–0.79]. Each range corresponds to one of the three experts generated through the evolution process.  Including the original trainable expert, this results in a total of four MoE experts.

\begin{table}[ht]
\centering
\caption{\textbf{Comparison of MLLMs on image understanding benchmarks.} `LLM' is the language model component, `Act.' is the number of activated parameters, and `Res.' is the input image resolution. Models `Q' ,`Q$^\prime$', `S', `P',  `ML' , `G', and `O' refer to Qwen~\cite{bai2023qwen}, Qwen2~\cite{yang2024qwen2}, StableLM~\cite{bellagente2024stable}, Phi-2~\cite{javaheripi2023phi}, Mobile LLaMA~\cite{kan2024mobile}, Gemini~\cite{team2024gemini} and OpenChat~\cite{wang2023openchat}, respectively. `AVG' is the weighted mean across all benchmarks, with MME values divided by 20 for simplify calculation. $^*$ 
indicates results re-implemented using MoE-LLaVA~\cite{lin2024moe}. Rows are colored based on the same baseline settings as our method for easier comparison.}
\label{tab:table1}
\begin{adjustbox}{max width=\textwidth}
\begin{tabular}{l|ccc|cccc|ccc|c}
\hline
                                   &                                &                                 &                                 & \multicolumn{4}{c|}{\textbf{Image Question Answering}}        & \multicolumn{3}{c|}{\textbf{Benchmark Toolkit}} &                                \\
\multirow{-2}{*}{\textbf{Methods}} & \multirow{-2}{*}{\textbf{LLM}} & \multirow{-2}{*}{\textbf{Act.}} & \multirow{-2}{*}{\textbf{Res.}} & VQA$^{v2}$         & GQA           & SQA           & VQA$^t$           & POPE          & MME             & MMB           & \multirow{-2}{*}{\textbf{AVG}} \\ \hline
\rowcolor[HTML]{F7F7F7} 
\textit{\textbf{0-1B}}             &                                &                                 &                                 &               &               &               &               &               &                 &               &                                \\
\rowcolor[HTML]{F7F7F7} 
\textit{\textbf{Sparse Model}}     &                                &                                 &                                 &               &               &               &               &               &                 &               &                                \\
\rowcolor[HTML]{ECF4FF} 
MoE-LLaVA$^*$                         & Q$^\prime$-0.5B                         & 0.6B                            & 336                             & \underline{72.0}          & \underline{56.1}          & \underline{58.0}          & \underline{39.6}          & \underline{84.4}          & \underline{1170.1}          & \underline{57.8}          & \underline{60.9}                           \\
\rowcolor[HTML]{F8E1D7} 
\textbf{EvoMoE}                    & Q$^\prime$-0.5B                         & 0.7B                            & 336                             & \textbf{74.4} & \textbf{57.4} & \textbf{59.1} & \textbf{42.4} & \textbf{85.0} & \textbf{1188.6} & \textbf{58.2} & \textbf{62.3}                  \\ \hline
\rowcolor[HTML]{F7F7F7} 
\textit{\textbf{1-2B}}             &                                &                                 &                                 &               &               &               &               &               &                 &               &                                \\
\rowcolor[HTML]{F7F7F7} 
\textit{\textbf{Sparse Model}}     &                                &                                 &                                 &               &               &               &               &               &                 &               &                                \\
\rowcolor[HTML]{ECF4FF} 
MoE-LLaVA~\cite{lin2024moe}                          & S-1.6B                         & 2.0B                            & 336                             & \underline{76.7}          & 60.3          & 62.6          & \underline{50.1}          & 85.7          & \underline{1318.2}          & {60.2}          & {65.9}                           \\
\rowcolor[HTML]{F8E1D7} 
\textbf{EvoMoE}&   S-1.6B                         & 1.8B                            & 336                             & \textbf{76.9}          & \underline{61.2}          & \textbf{63.5}          & \textbf{51.5}          & 86.4          & \textbf{1359.7}          & \underline{60.9 }          & \textbf{67.0}          \\
MoE-LLaVA~\cite{lin2024moe}                           & Q-1.8B                         & 2.2B                            & 336                             & 76.2          & \textbf{61.5}          & {63.1}          & 48.0          & \underline{87.0}          & 1281.6          & 59.7          & 65.7                           \\
\rowcolor[HTML]{ECF4FF} 
MoE-LLaVA$^*$                         & Q-1.8B                         & 2.2B                            & 336                             & 76.2          & 61.0          & 62.6          & 48.0          & 86.5          & 1288.1          & 59.4          & 65.3                           \\
\rowcolor[HTML]{F8E1D7} 
\textbf{EvoMoE}                    & Q-1.8B                         & 2.0B                              & 336                             & \textbf{76.9} & \underline{61.2} & \underline{63.3} & {49.3} & \textbf{87.1} & {1315.6} & \textbf{61.6} & \underline{66.5}                  \\ \hline
\rowcolor[HTML]{F7F7F7} 
\textit{\textbf{2-3B}}             &                                &                                 &                                 &               &               &               &               &               &                 &               &                                \\
\rowcolor[HTML]{F7F7F7} 
\textit{\textbf{Dense Model}}      &                                &                                 &                                 &               &               &               &               &               &                 &               &                                \\
TinyGPT-V~\cite{yuan2024tinygptvefficientmultimodallarge}                          & P-2.7B                         & 2.7B                            & 448                             & -             & 33.6          & 41.2         & 11.4          & 50.5          & 507.8           & 35.5          & -                              \\
Mini-Gemini~\cite{li2024minigeminiminingpotentialmultimodality}                        & G-2B                           & 2.0B                              & 336                             & -             & -             & -             & 56.2          & -             & 1341.0          & 59.8          & -                              \\
MobileVLM~\cite{chu2023mobilevlmfaststrong}                          & ML-2.7B                         & 2.7B                            & 336                             & -             & \textbf{85.4}          & 59.0          & 46.7          & 84.6          & 1296.4          & 57.0          & -                              \\
MobileVLM v2~\cite{chu2024mobilevlmv2fasterstronger}                       & ML-2.7B                         & 2.7B                            & 336                             & -             & 61.1          & 70.0          & 57.5          & 84.7          & 1440.5          & 63.2          & -                              \\
LLaVA-Phi~\cite{chu2024mobilevlmv2fasterstronger}                          & P-2.7B                         & 2.7B                            & 336                             & 71.4          & 68.4          & 66.4          & 48.6          & 85.0          & 1335.1          & 59.8          & 66.6                           \\
\rowcolor[HTML]{F7F7F7} 
\textit{\textbf{Sparse Model}}     &                                &                                 &                                 &               &               &               &               &               &                 &               &                                \\
Qwen-MoE$^*$~\cite{yang2024qwen2technicalreport}                          & P-2.7B                         & 2.7B                            & 336                             & 77.5          & 61.1          & 67.7          & 52.6          & 85.9          & 1434.0          & 65.4          & 68.9                           \\
\rowcolor[HTML]{ECF4FF} 
MoE-LLaVA~\cite{lin2024moe}                           & P-2.7B                         & 3.6B                            & 336                             & 77.6          & 61.4          & 68.5          & 51.4          & 86.3          & 1423.0          & 65.2          & 68.7                           \\
\rowcolor[HTML]{F8E1D7} 
\textbf{EvoMoE}                    & P-2.7B                         & 3.0B                              & 336                             & {77.8} & {61.6} & {69.5} & {52.0} & \textbf{86.6} & \textbf{1450.5} & {66.8} & {69.6}                  \\
\rowcolor[HTML]{ECF4FF} 
MoE-LLaVA~\cite{lin2024moe}                             & P-2.7B                         & 3.6B                            & 384                             & \underline{79.9}          & {62.6}          & \underline{70.3}          & \underline{57.0}          & 85.7          & 1431.3          & \underline{68.0}          & \underline{70.5}                           \\
\rowcolor[HTML]{F8E1D7} 
\textbf{EvoMoE}                    & P-2.7B                         & 3.0B                              & 384                             & \textbf{80.2} & \underline{62.8} & \textbf{71.5} & \textbf{57.8} & \underline{86.5} & \underline{1450.1} & \textbf{69.6} & \textbf{71.6}                  \\ \hline
\rowcolor[HTML]{EFEFEF} 
\textit{\textbf{7B}}               &                                &                                 &                                 &               &               &               &               &               &                 &               &                                \\
\rowcolor[HTML]{EFEFEF} 
\textit{\textbf{Sparse Model}}     &                                &                                 &                                 &               &               &               &               &               &                 &               &                                \\
\rowcolor[HTML]{ECF4FF} 
MoE-LLaVA$^*$                          & O-7B                           & 9.6B                            & 336                             & \underline{78.1}          & \underline{61.5}          & \underline{62.8}          & \underline{52.7}          & \underline{86.8}          & \underline{1384.5}          & \underline{64.8}          & \underline{67.9}                           \\
\rowcolor[HTML]{F8E1D7} 
\textbf{EvoMoE}                    & O-7B                           & 7.3B                            & 336                             & \textbf{78.9} & \textbf{62.6} & \textbf{63.8} & \textbf{53.8} & \textbf{87.3} & \textbf{1391.5} & \textbf{65.8} & \textbf{68.8}                  \\ \hline
\end{tabular}
\end{adjustbox}
\end{table}

\subsection{Comparison with State-of-the-Art}
\label{sec:Comparison}

We evaluated our method against state-of-the-art approaches on four image question-answering benchmarks and three multi-modal understanding toolkits. As illustrated in Table~\ref{tab:table1}, the models were categorized according to the size of  LLM into four groups: 0–1B, 1–2B, 2–3B, and 7B. 

Compared with the state-of-the-art method MoE-LLaVA, which serves as a baseline for MoE-tuning, EvoMoE demonstrates strong multi-modal understanding capabilities across various LLM sizes and image resolutions. EvoMoE outperforms MoE-LLaVA in the LLMs Qwen2-0.5B, StableLM-1.6B, Qwen-1.8B, Phi-2.7B, and OpenChat-7B. It achieves average performance gains of 1.4\% for the 0.5B model, 1.1\% for the 1.6B model, 1.2\% for the 1.8B model, 1.1\% for the 2.7B model and 0.9\% for the 7B model, all with fewer activated parameters (activating only top-1 expert). In particular, EvoMoE achieves remarkable improvements in the TextVQA, VQAv2, and GQA benchmarks. For instance, with the Qwen2-0.5B model, it surpasses the baseline by 2.8\%, 2.4\%, and 1.3\%, respectively. In StableLM-1.6B, EvoMoE improves TextVQA performance by 1.4\%. Additionally, it outperforms baselines in MMbench evaluations by 2.2\%, 1.6\%, and 1.0\% with Qwen-1.8B, Phi-2.7B, and OpenChat-7B models, respectively. 
This is particularly noteworthy, given that the baseline approach relies on activating the top-2 experts, which results in a significantly higher number of activated parameters. 
Ultimately, under the same Phi-2.7 LLM, EvoMoE outperformed the baseline by both 1.6\% on input image resolutions of both 336 and 384, demonstrating the flexibility of EvoMoE.
Collectively, these results demonstrate that EvoMoE not only outperforms other sparse models but also achieves this with fewer activated parameters.

\subsection{Comprehensive Analysis}
\label{sec:Ablation}

In this section, we peform an  ablation study to explore  EvoMoE's core contributions using the Qwen-1.8B model. We conduct experiments on four image QA benchmarks and three multi-modal understanding benchmarks, using the same training data as \cite{lin2024moe} for fair comparison.

\textbf{Design analysis of our framework.} We conduct several ablation studies to assess the effectiveness of the proposed framework.
As depicted in Table~\ref{tab:table2}, (a) represents a dense LLM without any MoE experts. (b) is a standard MoE model proposed by MoE-LLaVA~\cite{lin2024moe} based on MoE-tuning, which includes four experts and a linear router.
These results indicate that traditional MoE-tuning method does not provide a significant performance enhancement over dense LLMs on average accuracy. This is due to two challenges faced by MoE-tuning: expert uniformity and router rigidity.
In (c), we replaced the linear router with our proposed DTR router based on the MoE-tuning approach. The results indicate that DTR addresses the issue of router rigidity, thereby enhancing performance.
%
%
In (d), we tested our expert evolution strategy combined with a linear router. All newly created experts originate from the same dense LLM, which can be compared to (a).
The results demonstrate that our expert evolution strategy significantly enhances performance while maintaining a model size comparable to dense LLMs, effectively addressing the issue of expert uniformity.
Combined with the DTR router, our framework achieves optimal performance.

\begin{table}[!t]
    \centering
    \tablefontsizeone
    \caption{Ablation study on MLLM evaluation benchmarks.}
    \label{tab:table2}
    
    \begin{adjustbox}{max width=\textwidth} 
        \begin{tabular}{c|cccc|cccc|ccc|c}
        \hline
                   &                                &                                &                                &                                 & \multicolumn{4}{c|}{\textbf{Image Question Answering}}        & \multicolumn{3}{c|}{\textbf{Benchmark Toolkit}} &                       \\
            \multirow{-2}{*}{} & \multirow{-2}{*}{\textbf{ M-T\cite{lin2024moe}.}} & \multirow{-2}{*}{\textbf{Evo.}} & \multirow{-2}{*}{\textbf{DTR}} & \multirow{-2}{*}{\textbf{Act.}} & VQA$^{v2}$         & GQA           & SQA           & VQA$^{T}$          & POPE          & MME             & MMB           & \multirow{-2}{*}{\textbf{AVG}} \\ \cline{1-13}
            (a)                &                                &                                &                                & 1.8B                            & 76.3          & 61.0          & 62.1          & 48.2          & 86.4          & 1286.7          & 59.7          & 65.4                  \\
            (b)                & $\checkmark$                              &                                &                                & 2.2B                            & 76.2          & 61.0          & 62.6          & 48.0          & 86.5          & 1288.1          & 59.4          & 65.5                  \\
            (c)                & $\checkmark$                              &                                & $\checkmark$                              & 2.4B                            & 76.2          & \underline{61.1}          & \underline{63.0}          & 48.6          & \underline{86.8}          & 1310.4          & \underline{61.4}          & 66.0                  \\ \hline
            (d)                &                               & $\checkmark$                              &                                & \textbf{1.8B}                   & \textbf{77.5} & \textbf{61.2}          & 62.9          & \underline{48.8}          & \underline{86.8}          & \underline{1311.4}          & 61.3          & \underline{66.3}                  \\
            \rowcolor[HTML]{FFFFFF} 
            \textbf{EvoMoE}                &                     & \textbf{$\checkmark$}                     & \textbf{$\checkmark$}                     & 2.0B                            & \underline{76.9}          & \textbf{61.2} & \textbf{63.3} & \textbf{49.3} & \textbf{87.1} & \textbf{1315.6} & \textbf{61.6} & \textbf{66.5}         \\ \hline
        \end{tabular}
    \end{adjustbox}
\end{table}

\begin{table}[htbp]
    \centering
    \caption{Ablation study for evolution strategy.}
    \label{tab:table3}
    \footnotesize 
    \begin{tabular}{c|c|cccc|ccc}
    \hline
     & $\beta$ & VQA$^{v2}$         & GQA           & SQA           & VQA$^{T}$          & POPE          & MME             & MMB           \\ \hline
    Expert 1    & 1.0   & 76.3          & \underline{61.0} & 62.1          & 48.2          & \underline{86.4}          & 1286.7          & \underline{59.7}          \\
    Expert 2    & 0.9 & 76.4          & 60.8          & \underline{62.7}          & 48.6          & \textbf{87.3} & \textbf{1305.7} & 58.4          \\
    Expert 3    & 0.8 & \underline{76.7}          & 60.9          & 62.4          & \textbf{49.0} & 86.6          & \underline{1297.3}          & \textbf{61.4} \\
        Expert 4    & 0.7 & \textbf{77.1} & \textbf{61.2}          & \textbf{62.8} & \underline{48.7}          & \underline{86.4}          & 1284.5          & 59.5          \\ 
        \hline
    \end{tabular}
\end{table}

\textbf{The Effectiveness of Evolution Strategy.} 
Table~\ref{tab:table3} demonstrates the potential of our proposed expert evolution strategy in MLLMs. In this experiment, we eliminated the router and independently evaluated the multi-modal capabilities of each expert. By fixing $\beta$ to a constant value, we systematically examined its impact on performance.
Expert 1 is a FFN layer with $\beta=1.0$, while Experts 2 to 4 are evolved from Expert 1 by progressively reducing $\beta$ values.
Notably, the evolved experts consistently outperform Expert 1 across the majority of benchmarks. This performance advantage persists even under large $\beta$ decay, where the value is set to 0.9, equivalent to retaining only 10\% of updates per step. 
These systematic improvements empirically validate our core hypothesis: experts generated through expert evolution exhibit significantly greater diversity compared to those generated through straightforward replication. Each evolved expert demonstrates specialized capabilities, excelling in different benchmarks, often outperforming the original expert and effectively addressing the challenge of expert uniformity.
In the subsequent stage, we apply the proposed DTR to these evolved experts, enabling better utilization of their specialized capabilities and enhancing overall performance.
In our experiments, to enhance generalization, we randomly sample the $\beta$ value from a predefined range at each training step, rather than using a fixed value.

\begin{table}[htbp]
    \begin{minipage}[t]{0.5\textwidth}  
        \caption{Ablation study for DTR.} 
        \label{tab:table4}
        \tablefontsizethree
        \raggedright                  
        \resizebox{1.0\linewidth}{!}{    
        \begin{tabular}{cccc|cccc}
        \hline
                 & \textbf{Share} & \textbf{Image}      & \textbf{Text}       & \textbf{GQA}           & \textbf{SQA}           & \textbf{VQA$^t$}          & \textbf{POPE}          \\ \hline
        \textit{\textbf{Linear}}   &         &            &            &               &               &               &               \\
            (a)  & $\checkmark$        &            &            & 61.0          & 62.6          & 48.0          & 86.5          \\
             (b)    &         & $\checkmark$           & $\checkmark$           & \textbf{61.2}          & \underline{62.7}          & 48.3          & 86.6          \\
           (c)      & $\checkmark$        & $\checkmark$           & $\checkmark$           & \underline{61.1}          & 62.2          & 48.2          & 86.4          \\ \hline
        \textit{\textbf{HyperNet}} &         &            &            &               &               &               &               \\
            DTR     &         & \textbf{$\checkmark$ } & \textbf{$\checkmark$ } & \textbf{{61.2}} & \textbf{63.3} & \textbf{49.2} & \textbf{87.1} \\
            (d)     & $\checkmark$        & $\checkmark$           & $\checkmark$           & 60.9          & \underline{62.7}          & \underline{48.4}          & \underline{86.7}          \\ \hline
        \end{tabular}
        }
    \end{minipage}%
    \begin{minipage}[t]{0.5\textwidth}  
        \caption{Ablation study for expert diversity.} 
        \label{tab:table5}
        \raggedright                  
        \resizebox{1.0\linewidth}{!}{    
        \begin{tabular}{c|ccccc}
        \hline
                            & \multicolumn{1}{c|}{\textbf{Method}}    & \textbf{GQA}           & \textbf{SQA}           & \textbf{VQA$^t$}          & \textbf{POPE}          \\ \hline
        (a)                     & \multicolumn{1}{c|}{}            & 61.0          & 62.6          & 48.0          & 86.5          \\
        \textit{\textbf{Initialization}} & \multicolumn{1}{c|}{\textit{}}   &               &               &               &               \\
        (b)                     & \multicolumn{1}{c|}{Noise}       & 60.8          & \underline{63.1}          & 47.2          & 86.1          \\
        (c)                     & \multicolumn{1}{c|}{V-Evo.}         & \underline{61.3}          & {63.0} & 48.0          & \underline{86.7} \\
        \textit{\textbf{Training}}       & \multicolumn{1}{c|}{\textit{}}                    &               &               &               &               \\
        (d)                     & \multicolumn{1}{c|}{Dropout}     & 60.6          & 62.3          & 47.4          & 86.2          \\
        (e)                     & \multicolumn{1}{c|}{Contrastive} & \textbf{61.5} & 62.6          & 47.5          & 86.3          \\
        (f)                     & \multicolumn{1}{c|}{Local Loss}   & 60.9          & 62.2          & \underline{48.2} & 85.9          \\ 
        \textbf{Ours}                     & \multicolumn{1}{c|}{\textbf{Evolution}}   & {61.2}          & \textbf{63.3}          & \textbf{49.3} & \textbf{87.1}          \\ 
        \hline
        \end{tabular}
        }
    \end{minipage}
\end{table}

\begin{wrapfigure}{r}{0.45\textwidth}
  \vspace{-20pt}
  \begin{center}
    \includegraphics[width=0.45\textwidth]{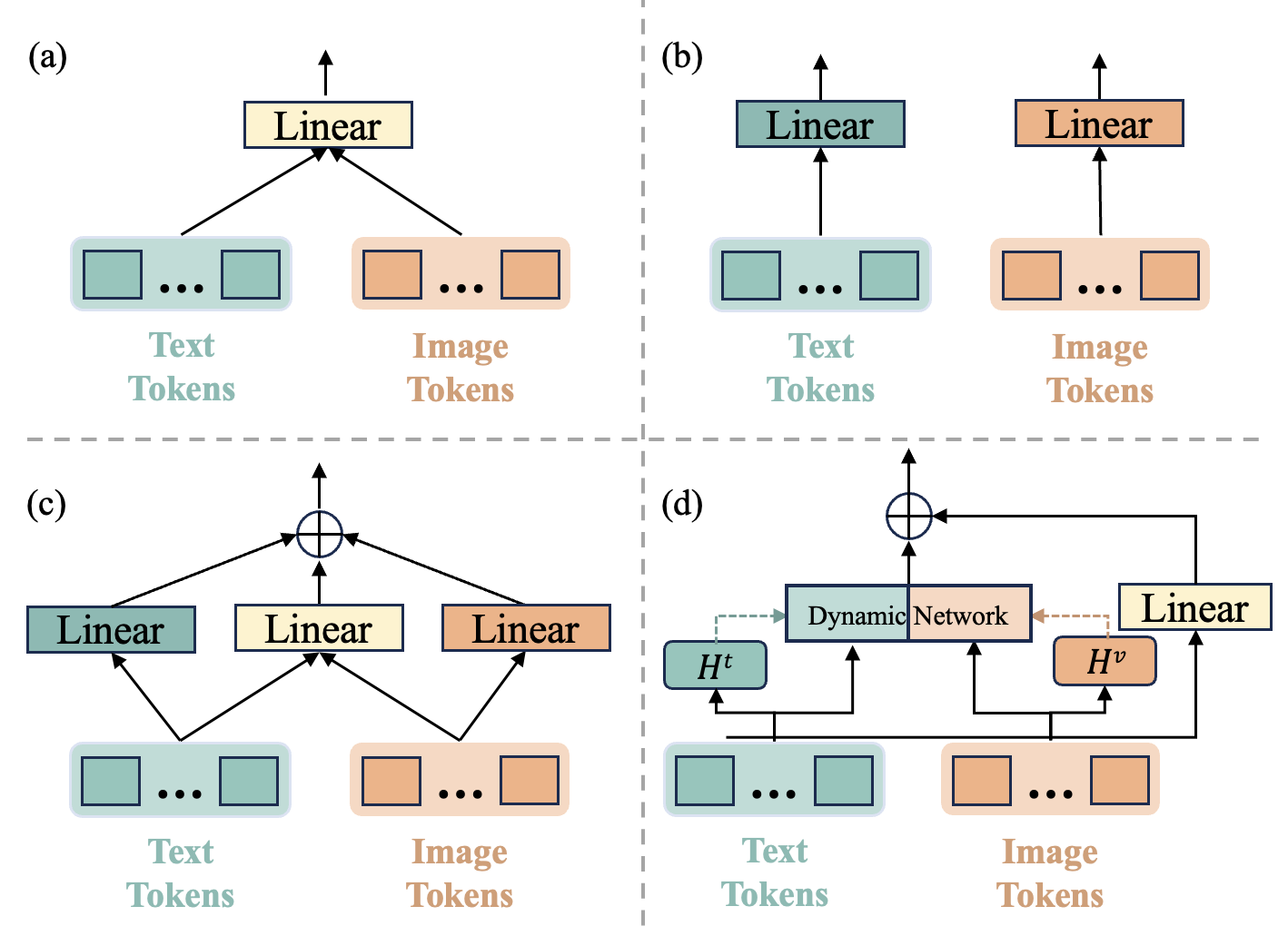}
  \end{center}
  \caption{\textbf{Design analysis of DTR.} (a) single router; (b) modality-specific router; (c) modality-specific router with shared routing; (d) hyperNet with shared routing.}
  \vspace{-10pt}
  \label{figure:DTR}
\end{wrapfigure}

\textbf{Design analysis of DTR.} Figure~\ref{figure:DTR} presents possible architectures for DTR. (a) presents a standard linear router that processes image and text tokens simultaneously. (b) introduces a modality-specific router tailored to differentiate between image and text. (c) incorporates a shared router, which influences the modality-specific router through weighted connections. (d) proposes the hypernetwork as the modality-specific router while also integrating a shared router to enhance flexibility.

\begin{figure}[h!]
    \centering
    \begin{subfigure}{0.470\textwidth}
        \centering
        \includegraphics[width=\linewidth]{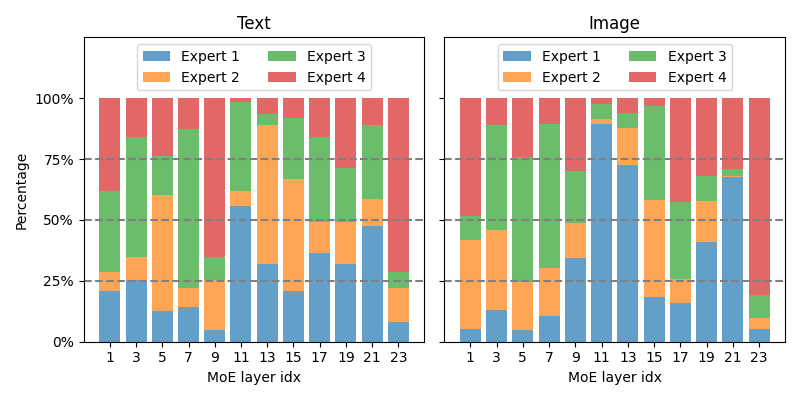} 
        \caption{MoE-LLaVA [4]} 
        \label{fig:image1}
    \end{subfigure}
    \hfill
    \begin{subfigure}{0.47\textwidth}
        \centering
        \includegraphics[width=\linewidth]{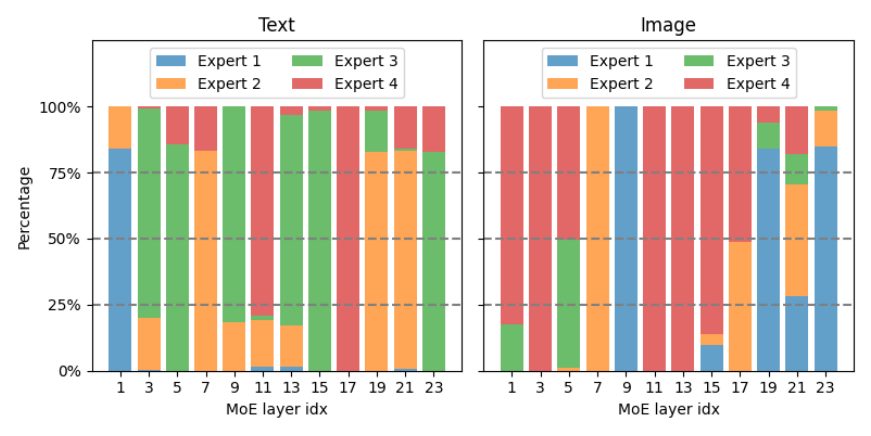} 
        \caption{EvoMoE} 
        \label{fig:image2}
    \end{subfigure}
   \caption{\textbf{Distribution of input modalities across different experts} (a) Previous methods exhibited almost uniform distributions across different inputs, leading to router rigidity.
(b) EvoMoE dynamically allocates input tokens to the most suitable experts based on their modality.}
    \label{fig:router_vis}
\end{figure}

Table~\ref{tab:table4} summarizes the ablation studies. The modality-specific router in (b) outperforms the single router in (a), emphasizing the importance of modality distinction. The HyperNet adaptation improves attention to input token distribution, further improving performance. However, adding a weighted shared router in (c) and (d) results in a decline in overall performance. Ultimately, we adopt the structure of the DTR in Figure~\ref{figure:DTR_Evo} in our framework, as it achieves the best performance.
Figure~\ref{fig:router_vis} visualizes modality preferences of experts on the ScienceQA benchmark, with MoE-tuning results on the left and EvoMoE results on the right. 
%
%
The visualization reveals that traditional MoE-tuning exhibits almost uniform distributions across different inputs, leading to router rigidity. In contrast, EvoMoE, using DTR, dynamically allocates tokens to suitable experts based on modality, allowing experts to learn specific patterns for efficient, input-guided processing.


\textbf{Increasing Expert Diversity.} 
To address homogenization from expert replication, we implemented strategies to improve expert diversity, classified into initialization and training phases. As shown in Table~\ref{tab:table5}, (a) shows the MoE baseline. (b) adds noise during expert initialization, while (c) Vanilla-Evolution shifts expert evolution to Stage I and fine-tunes all experts in Stage II.
For training: (d) uses random dropout; (e) incorporates NCE loss~\cite{chen2020simple} among experts; and (f) introduces local loss~\cite{mustafa2022multimodal} to increase router entropy for better routing balance. These diversity strategies didn't significantly boost performance across all metrics, highlighting EvoMoE's superiority. For detailed comparison, see the Supplementary Material.


\textbf{MoE Strategy Exploration.} 
We further explored additional attempts concerning MoE in Table~\ref{tab:table6}. 
Incorporating insights from advanced LLMs like DeepSeek-V3~\cite{liu2024deepseek}, it was found that removing the initial MoE layer, emphasized in DeepSeek-V3, is ineffective in MLLMs. While shared experts are common in LLM MoE implementations, they have not provided significant benefits in MLLMs.
%
%
Additionally, we explored the introduction of additional trainable parameters at various stages:
(1) In stage II, unfreezing all parameters (MSA\&FFN) within the LLM led to optimal performance on several benchmarks (GQA, SQA, MMB), though improvements were not consistent across all benchmarks. 
(2) In stage III, training the entire set of experts alongside the DTR led to a significant performance drop.
Lastly, our framework preserved performance using an alternating approach for MoE layers, whereas replacing all dense layers with MoE structures decreased performance.



\begin{table}[htbp]
\centering 
\caption{Ablation study on MoE exploration.} 
\label{tab:table6}
\tablefontsizetwo
\begin{tabular}{c|cccc|ccc}
\hline
\textbf{Strategy}                  & \textbf{VQA$^{v2}$} & \textbf{GQA}  & \textbf{SQA} & \textbf{VQA$^{t}$} & \textbf{POPE} & \textbf{MME} & \textbf{MMB} \\ \hline
\rowcolor[HTML]{EFEFEF} 
\textit{\textbf{LLM MoE Insights}}                   &                &               &              &                &               &              &              \\ 
\rowcolor[HTML]{FFFFFF} 
w/o first layer                    & 76.3           & \underline{61.1}          & 62.4         & 48.5           & 86.4          & 1285.6       & 60.6         \\ 
\rowcolor[HTML]{FFFFFF} 
Share Expert                       & 76.2           & 61.0          & 62.6         & \underline{48.8}           & \underline{86.6}          & \underline{1306.8}       & 60.8         \\ \hline
\rowcolor[HTML]{EFEFEF} 
\textit{\textbf{Trainable Parameters}} &                &               &              &                &               &              &              \\ 
\rowcolor[HTML]{FFFFFF} 
In Stage II                           & 75.2           & \textbf{61.2} & \textbf{63.8}& 46.8           & 86.4          & 1263.8       & \textbf{61.9} \\ 
\rowcolor[HTML]{FFFFFF} 
In Stage III                          & \textbf{77.0}           & 60.9          & 62.3         & 48.4           & 86.5          & 1271.2       & 60.9         \\ \hline
\rowcolor[HTML]{EFEFEF} 
\textit{\textbf{MoE Placement}}    &                &               &              &                &               &              &              \\ 
\rowcolor[HTML]{FFFFFF} 
ALL Layers                              & 74.4           & 61.0          & 62.5         & 47.6           & 86.3          & 1280.1       & 60.4         \\ 
\rowcolor[HTML]{FFFFFF} 
\textbf{EvoMoE}                             & \underline{76.9}           & \textbf{61.2} & \underline{63.3}         & \textbf{49.3}  & \textbf{87.1} & \textbf{1315.6} & \underline{61.6}         \\ \hline
\end{tabular}
\end{table}

\section{Conclusion}
\label{sec:conclusion}

In this paper, we introduce EvoMoE, a MoE framework specifically designed for MLLMs. EvoMoE redefines MoE-tuning through two key innovations: expert evolution and the dynamic token-aware router (DTR), effectively addressing two critical challenges in existing MoE-tuning approaches: expert uniformity and router rigidity. The superior performance of EvoMoE, validated through extensive experiments, highlights its potential to unlock new possibilities for the application of MoE in MLLMs.

{
\bibliographystyle{splncs04}
\bibliography{neurips_2025}
}


\newpage
\appendix

\begin{center}
    {\LARGE \textbf{Supplementary Materials}} \\[1em] 
\end{center}


\section{Overview}
\label{sec:oveeview}

This document provides a list of supplemental materials to support the main paper. It includes more ablation studies and visualization cases specifically based on the Qwen-1.8B model.

\section{Ablation Studies}

\subsection{Training Details.} Table~\ref{table:hyper} presents the training hyperparameters used across three stages for all models evaluated within our framework, including Qwen2-0.5B, StableLM-1.6B, Qwen-1.8B, Phi2-2.7B, and OpenChat-7B.

\begin{table}[htbp]
\centering
\caption{Training hyperparameters. }
\label{table:hyper}
\begin{tabular}{l|ccc}
\hline
\textbf{Configuration} & \textbf{Stage I} & \textbf{Stage II} & \textbf{Stage III} \\ \hline
Experts                & -                & -                 & 4                  \\
Top-k                  & -                & -                 & 1                  \\
Data                   & Hybird-PT        & LLaVA-FT          & LLaVA-FT           \\
Deepspeed              & Zero2            & Zero2             & Zero2\_offload     \\
Image Resolution       & \multicolumn{3}{c}{336*336}                               \\
Image encoder          & \multicolumn{3}{c}{CLIP-Large/336}                        \\
Image projector        & \multicolumn{3}{c}{MLP with GeLU}                         \\
Epoch                  & \multicolumn{3}{c}{1}                                     \\
Learning rate          & \multicolumn{3}{c}{2e-5}                                  \\
Learning rate schdule  & \multicolumn{3}{c}{Cosine decay}                          \\
Weight decay           & \multicolumn{3}{c}{0.0}                                   \\
Text max length        & \multicolumn{3}{c}{2048}                                  \\
Precision              & \multicolumn{3}{c}{Bf16}                                  \\
Global batch size      & 256              & 64                & 64                 \\
Training steps         & 5200             & 10395             & 10395              \\
Training hours         & 6.0              & 12.0              & 7.0                \\
Epoch                  & \multicolumn{3}{c}{1}                                     \\
GPU                    & \multicolumn{3}{c}{8xA100-80G}                            \\ \hline
\end{tabular}
\end{table}

\subsection{Training Objective.}
The overall loss function of EvoMoE consists of two components: the regression loss:  $\mathcal{L}_{\text {regressive }}$ and the auxiliary loss $\mathcal{L}_{\text {aux}}$. Regression loss is designed to optimize model performance, while auxiliary loss aims to promote a balanced load distribution across the router:

\begin{equation}
\mathcal{L}_{\text {total }}=\mathcal{L}_{\text {regressive }}+\alpha \cdot \mathcal{L}_{\text {aux }.}
\end{equation}
Here, $\alpha$ is a hyperparameter that controls the weight of the auxiliary loss and is set to 0.001 during the training process.

\textbf{Auto-Regressive Loss.} The output of EvoMoE is denoted by $\Upsilon$, which represents a sequence generated progressively, with each text element produced step-by-step:

\begin{equation}
\mathcal{L}_{\text {regressive }}=-\sum_{i=1}^N \log p\left(\Upsilon^{[i]} \mid \vartheta, \Gamma^{[: i-1]}\right)
\end{equation}

where $\vartheta$ denote the output of the vision embedding from the projection layer, $\Gamma$ represent the output of the text embedding from the word embedding layer. $N$ is the length of the output sequence.

\textbf{Balance Loss.} A differentiable load balance loss is employed in each router layer to encourage experts to process tokens in a balanced manner, as defined below:

\begin{equation}
\mathcal{L}_{\mathrm{aux}}=E \cdot \sum_{i=1}^E \mathcal{F}_i \cdot \mathcal{G}_i
\end{equation}

where $\mathcal{F}$ denotes the fraction of tokens processed by each expert. $\mathcal{G}$ epresents the average routing probability for each expert, and $E=4$ is the total number of experts in our paper.

\subsection{More experiments.}

\textbf{Architecture details of EvoMoE.}
Table~\ref{table:Architecture} details the activated and training parameters for the dense model, EvoMoE, and our baseline, MoE-LLaVA, across multiple LLMs. The results show that EvoMoE activates only the top-1 expert while adding a minimal number of parameters through Dynamic Token-aware Router (DTR). Consequently, the number of activated parameters in EvoMoE is lower than that of MoE-LLaVA. Notably, during training, EvoMoE updates only a single FFN, with additional experts generated through the evolution of this primary FFN. This approach leads to a significant reduction in the total parameter count compared to MoE-LLaVA, thereby enhancing overall efficiency.

\begin{table}[htbp]
\caption{Architecture details of EvoMoE. }
\label{table:Architecture}
\tablefontsizethree
\centering
\begin{adjustbox}{max width=\textwidth}
\begin{tabular}{l|ccc|ccccccc|cc}
\hline
\multirow{2}{*}{\textbf{Model}} & \multirow{2}{*}{\textbf{Experts}} & \multirow{2}{*}{\textbf{Top-k}} & \textbf{MoE}    & \multirow{2}{*}{\textbf{Embedding}} & \multirow{2}{*}{\textbf{Width}} & \multirow{2}{*}{\textbf{Layers}} & \multirow{2}{*}{\textbf{FFN}}   & \textbf{FFN}                & \multirow{2}{*}{\textbf{Heads}} & \multirow{2}{*}{\textbf{Router}} & \textbf{Activated} & \textbf{Training} \\
                       &                          &                        & \textbf{Layers} &                            &                        &                         &                        & \textbf{Factor}             &                        &                         & \textbf{Param.}     & \textbf{Param.}    \\ \hline
Qwen2-0.5B             & -                        & -                      & -      & \multirow{3}{*}{151936}    & \multirow{3}{*}{1024}  & \multirow{3}{*}{24}     & \multirow{3}{*}{2816}  & \multirow{3}{*}{3} & \multirow{3}{*}{24}    & -                       & 0.5B      & 0.5B     \\
MoE-LLaVA              & 4                        & 2                      & 12     &                            &                        &                         &                        &                    &                        & 2048                    & 0.6B      & 0.8B     \\
EvoMoE                 & 4                        & 1                      & 12     &                            &                        &                         &                        &                    &                        & 34760                   & 0.7B      & 0.7B     \\ \hline
StableLM-1.6B          & -                        & -                      & -      & \multirow{3}{*}{100352}    & \multirow{3}{*}{2560}  & \multirow{3}{*}{32}     & \multirow{3}{*}{10240} & \multirow{3}{*}{2} & \multirow{3}{*}{32}    & -                       & 1.6B      & 1.6B     \\
MoE-LLaVA              & 4                        & 2                      & 16     &                            &                        &                         &                        &                    &                        & 2048                    & 2.0B      & 2.9B     \\
EvoMoE                 & 4                        & 1                      & 16     &                            &                        &                         &                        &                    &                        & 34760                   & 1.8B      & 1.8B     \\ \hline
Qwen-1.8B              & -                        & -                      & -      & \multirow{3}{*}{151936}    & \multirow{3}{*}{2048}  & \multirow{3}{*}{24}     & \multirow{3}{*}{5504}  & \multirow{3}{*}{3} & \multirow{3}{*}{16}    & -                       & 1.8B      & 1.8B     \\
MoE-LLaVA              & 4                        & 2                      & 12     &                            &                        &                         &                        &                    &                        & 2048                    & 2.2B      & 3.1B     \\
EvoMoE                 & 4                        & 1                      & 12     &                            &                        &                         &                        &                    &                        & 34760                   & 2.0B      & 2.0B     \\ \hline
Phi2-2.7B              & -                        & -                      & -      & \multirow{3}{*}{51200}     & \multirow{3}{*}{2560}  & \multirow{3}{*}{32}     & \multirow{3}{*}{10240} & \multirow{3}{*}{2} & \multirow{3}{*}{32}    & -                       & 2.7B      & 2.7B     \\
MoE-LLaVA              & 4                        & 2                      & 16     &                            &                        &                         &                        &                    &                        & 2048                    & 3.6B      & 5.3B     \\
EvoMoE                 & 4                        & 1                      & 16     &                            &                        &                         &                        &                    &                        & 34760                   & 4.5B      & 7.8B     \\ \hline
OpenChat-7B            & -                        & -                      & -      & \multirow{3}{*}{32000}     & \multirow{3}{*}{4096}  & \multirow{3}{*}{32}     & \multirow{3}{*}{14366} & \multirow{3}{*}{3} & \multirow{3}{*}{32}    & -                       & 6.7B      & 6.7B     \\
MoE-LLaVA              & 4                        & 2                      & 16     &                            &                        &                         &                        &                    &                        & 2048                    & 9.6B      & 15.2B    \\
EvoMoE                 & 4                        & 1                      & 16     &                            &                        &                         &                        &                    &                        & 34760                   & 7.3B      & 7.3B     \\ \hline
\end{tabular}
\end{adjustbox}
\end{table}

\textbf{Performance Comparison of Different Model Sizes.} Table~\ref{table:Size} presents a comprehensive performance comparison of dense, MoE, and EvoMoE models across various LLM sizes. Utilizing EvoMoE significantly enhances the final performance of the LLMs.

\begin{table}[htbp]
\centering
\caption{Ablation study about the model size of EvoMoE.}
\label{table:Size}
\begin{adjustbox}{max width=\textwidth}
\begin{tabular}{l|c|cc|cccccccc}
\hline
\textbf{Model}                     & \textbf{Size}                  & \textbf{MoE}          & \textbf{Evo.}         & \textbf{VQA$^{v2}$}         & \textbf{GQA}           & \textbf{SQA}           & \textbf{VQA$^t$}          & \textbf{POPE}          & \textbf{MME}             & \textbf{MMB}           & \textbf{AVG}           \\ \hline
\multirow{3}{*}{Qwen2}    & \multirow{3}{*}{0.5B} & $\times$            & $\times$            & 71.8          & 56.1          & 57.7          & 39.7          & 84.3          & 1168.8          & 57.9          & 60.7          \\
                          &                       & $\checkmark$           & $\times$           & 72.0          & 56.1          & 58.0          & 39.6          & 84.4          & 1170.1          & 57.8          & 60.9          \\
                          &                       & \cellcolor[HTML]{FFCCC9}$\checkmark$ & \cellcolor[HTML]{FFCCC9}$\checkmark$ & \cellcolor[HTML]{FFCCC9}\textbf{74.4} & \cellcolor[HTML]{FFCCC9}\textbf{57.4} & \cellcolor[HTML]{FFCCC9}\textbf{59.1} & \cellcolor[HTML]{FFCCC9}\textbf{42.4} & \cellcolor[HTML]{FFCCC9}\textbf{85.0} & \cellcolor[HTML]{FFCCC9}\textbf{1188.6} & \cellcolor[HTML]{FFCCC9}\textbf{58.2} & \cellcolor[HTML]{FFCCC9}\textbf{62.3} \\ \hline
\multirow{3}{*}{StableLM} & \multirow{3}{*}{1.6B} & $\times$            & $\times$            & 76.6          & 60.1          & 62.5          & 50.1          & 85.2          & 1315.1          & 60.1          & 65.7          \\
                          &                       & $\checkmark$          & $\times$             & 76.7          & 60.3          & 62.6          & 50.1          & 85.7          & 1318.2          & 60.2          & 65.9          \\
                          &                       & \cellcolor[HTML]{FFCCC9}$\checkmark$ & \cellcolor[HTML]{FFCCC9}$\checkmark$ & \cellcolor[HTML]{FFCCC9}\textbf{76.9} & \cellcolor[HTML]{FFCCC9}\textbf{61.2} & \cellcolor[HTML]{FFCCC9}\textbf{63.5} & \cellcolor[HTML]{FFCCC9}\textbf{51.5} & \cellcolor[HTML]{FFCCC9}\textbf{86.4} & \cellcolor[HTML]{FFCCC9}\textbf{1359.7} & \cellcolor[HTML]{FFCCC9}\textbf{60.9} & \cellcolor[HTML]{FFCCC9}\textbf{67.0} \\ \hline
\multirow{3}{*}{Qwen}     & \multirow{3}{*}{1.8B} & $\times$             & $\times$             & 76.3          & 61.0          & 62.1          & 48.2          & 86.4          & 1286.7          & 59.7          & 65.4          \\
                          &                       & $\checkmark$          & $\times$             & 76.2          & 61.0          & 62.6          & 48.0          & 86.5          & 12881           & 59.4          & 65.5          \\
                          &                       & \cellcolor[HTML]{FFCCC9}$\checkmark$ & \cellcolor[HTML]{FFCCC9}$\checkmark$ & \cellcolor[HTML]{FFCCC9}\textbf{76.9} & \cellcolor[HTML]{FFCCC9}\textbf{61.2} & \cellcolor[HTML]{FFCCC9}\textbf{63.3} & \cellcolor[HTML]{FFCCC9}\textbf{49.3} & \cellcolor[HTML]{FFCCC9}\textbf{87.1} & \cellcolor[HTML]{FFCCC9}\textbf{1315.6} & \cellcolor[HTML]{FFCCC9}\textbf{61.6} & \cellcolor[HTML]{FFCCC9}\textbf{66.5} \\ \hline
\multirow{3}{*}{Phi-2}    & \multirow{3}{*}{2.7B} & $\times$             & $\times$             & 77.4          & 61.1          & 68.5          & 51.5          & 86.0          & 1418.1          & 65.1          & 68.5          \\
                          &                       & $\checkmark$          & $\times$             & 77.6          & 61.4          & 68.5          & 51.4          & 86.3          & 1423.0          & 65.2          & 68.7          \\
                          &                       & \cellcolor[HTML]{FFCCC9}$\checkmark$ & \cellcolor[HTML]{FFCCC9}$\checkmark$ & \cellcolor[HTML]{FFCCC9}\textbf{77.8} & \cellcolor[HTML]{FFCCC9}\textbf{61.6} & \cellcolor[HTML]{FFCCC9}\textbf{69.5} & \cellcolor[HTML]{FFCCC9}\textbf{52.0} & \cellcolor[HTML]{FFCCC9}\textbf{86.6} & \cellcolor[HTML]{FFCCC9}\textbf{1450.5} & \cellcolor[HTML]{FFCCC9}\textbf{66.8} & \cellcolor[HTML]{FFCCC9}\textbf{69.6} \\ \hline
\multirow{3}{*}{OpenChat} & \multirow{3}{*}{7B}   & $\times$             & $\times$             & 78.2          & 61.5          & 62.9          & 52.6          & 86.8          & 1355.5          & 65.1          & 67.8          \\
                          &                       & $\checkmark$          & $\times$             & 78.1          & 61.5          & 62.8          & 52.7          & 86.8          & 1384.5          & 64.8          & 67.9          \\
                          &                       & \cellcolor[HTML]{FFCCC9}$\checkmark$ & \cellcolor[HTML]{FFCCC9}$\checkmark$ & \cellcolor[HTML]{FFCCC9}\textbf{78.9} & \cellcolor[HTML]{FFCCC9}\textbf{62.6} & \cellcolor[HTML]{FFCCC9}\textbf{63.8} & \cellcolor[HTML]{FFCCC9}\textbf{53.8} & \cellcolor[HTML]{FFCCC9}\textbf{87.3} & \cellcolor[HTML]{FFCCC9}\textbf{1391.5} & \cellcolor[HTML]{FFCCC9}\textbf{65.8} & \cellcolor[HTML]{FFCCC9}\textbf{68.8} \\ \hline
\end{tabular}
\end{adjustbox}
\end{table}

\textbf{Shuffle Router in MoE-tuning.} In Table~\ref{table:shuffle}, we conducted multiple tests to evaluate the impact of the shuffle router on MoE-tuning. MoE tuning was performed using replicate initialization, serving as a supplement to Figure 1(a) in the paper. The results demonstrate that the shuffle router does not affect the overall average performance. This finding confirms our hypothesis that traditional MoE-tuning methods suffer from significant expert homogeneity issues. Consequently, randomly selecting different routers makes no substantial difference, a phenomenon we refer to as \textbf{expert uniformity}.

\begin{table}[htbp]
\centering
\caption{Shuffle Router in MoE-tuning. }
\label{table:shuffle}
\begin{tabular}{c|cccc|ccc|c}
\hline
                                   & \multicolumn{4}{c|}{\textbf{Image Question Answering}} & \multicolumn{3}{c|}{\textbf{Benchmark Toolkit}} &                       \\
\multirow{-2}{*}{\textbf{Methods}} & VQA$^{v2}$        & GQA         & SQA         & VQA$^t$         & POPE          & MME             & MMB           & \multirow{-2}{*}{\textbf{AVG}} \\ \hline
MoE-tuning                        & 76.2         & 61.0        & 62.6        & 48.0        & 86.5          & 1288.1          & 59.4          & 65.5                  \\ \hline
 \textit{\textbf{Shuffle Router}}                     &       &      &        &       &        &        &       &  \\
1                     & 76.2         & 60.0        & 62.3        & 48.2        & 86.4          & 1288.5          & 59.6          & 65.3                  \\ 
2                     & 76.2         & 60.1        & 62.1        & 48.2        & 86.3          & 1288.8          & 59.4         & 65.2                  \\ 
3                     & 76.2         & 60.4        & 62.6        & 47.7        & 86.5          & 1290.2          & 59.8          & 65.4                 \\ 
4                     & 76.3         & 60.6        & 62.5        & 48.3        & 86.4          & 1293.2          & 59.5          & 65.6                  \\ 
5                     & 76.1         & 60.9        & 62.6        & 47.9        & 86.5          & 1289.9          & 59.3          & 65.5             \\ 
6                     & 76.2         & 60.6        & 62.5        & 47.8        & 86.4          & 1287.8          & 59.6          & 65.5               \\ 
7                    & 76.1         & 60.7        & 62.6        & 47.9        & 86.3          & 1286.9          & 59.3          & 65.4   \\
8                    & 76.0         & 60.1       & 62.2       & 47.9        & 86.4          & 12887.1          & 59.2         & 65.2   \\

\hline
\end{tabular}
\end{table}

\textbf{Training setting.} Tables~\ref{table:top-k},~\ref{table:expert_number}, and~\ref{table:epoch} outline the training settings for EvoMoE. To evaluate the \textbf{effect of the number of activated experts,} we compare the performance using different top-k strategies. As shown in Table~\ref{table:top-k}, our method demonstrates that top-1 experts performs significantly better than top-2, which is contrary to previous MoE experimental results, thereby confirming the greater efficiency of our approach. To verify the \textbf{impact of the number of experts} on the results, we conducted experiments presented in Table~\ref{table:expert_number}. Increasing the number of experts slightly enhances performance, validating previous findings that more sparse experts can achieve better results. Finally, we examine the influence of \textbf{training epochs}. Table~\ref{table:epoch} shows that when training for 2 epochs, performance on GQA increases significantly, while other metrics experience varying degrees of decline. This indicates that the network tends to overfit on large-scale datasets.

\begin{table}[htbp]
\caption{The value of top-k. }
\label{table:top-k}
\centering
\begin{adjustbox}{max width=\textwidth}
\begin{tabular}{c|ccccccc}
\hline
\textbf{Top-k}      & \textbf{VQA$^{v2}$}         & \textbf{GQA}           & \textbf{SQA}           & \textbf{VQA$^t$}          & \textbf{POPE}          & \textbf{MME}             & \textbf{MMB}           \\ \hline
\textbf{1} & \textbf{76.9} & \textbf{61.2} & \textbf{63.3} & \textbf{49.3} & \textbf{87.1} & \textbf{1315.6} & \textbf{61.6} \\
2          & 76.6          & 61.0          & 62.4          & 48.8          & 86.8          & 1304.7          & 60.9          \\ \hline
\end{tabular}
\end{adjustbox}
\end{table}

\begin{table}[htbp]
\caption{The number of experts. }
\label{table:expert_number}
\centering
\begin{adjustbox}{max width=\textwidth}
\begin{tabular}{c|ccccccc}
\hline
\textbf{Experts}      & \textbf{VQA$^{v2}$}         & \textbf{GQA}           & \textbf{SQA}           & \textbf{VQA$^t$}          & \textbf{POPE}          & \textbf{MME}             & \textbf{MMB}           \\ \hline
2          & 76.5          & 61.1          & 62.5          & 48.5          & 86.6          & 1302.6          & 61.3          \\
\textbf{4} & \textbf{76.9} & \textbf{61.2} & \textbf{63.3} & \textbf{49.3} & \textbf{87.1} & \textbf{1315.6} & \textbf{61.6} \\ \hline
\end{tabular}
\end{adjustbox}
\end{table}


\begin{table}[h!]
\caption{Training Epochs. }
\label{table:epoch}
\centering
\begin{adjustbox}{max width=\textwidth}
\begin{tabular}{c|ccccccc}
\hline
\multicolumn{1}{c|}{\textbf{Epoch}}      & \textbf{VQA$^{v2}$}         & \textbf{GQA}           & \textbf{SQA}           & \textbf{VQA$^t$}          & \textbf{POPE}          & \textbf{MME}             & \textbf{MMB}           \\ \hline
\multicolumn{1}{c|}{\textbf{1}} & \textbf{76.9} & 61.2          & \textbf{63.3} & \textbf{49.3} & \textbf{87.1} & \textbf{1315.6} & \textbf{61.6} \\ 
2                               & 76.2          & \textbf{62.4} & 61.8          & 48.4          & 86.5          & 1262.2          & 61.0          \\ \hline
\end{tabular}
\end{adjustbox}
\end{table}



\textbf{Details in Increasing Expert Diversity.} Figure~\ref{table:noise_dropout} illustrates the experimental details related to noise and dropout as discussed in the "Increasing Expert Diversity" section of the paper. We introduced various types of noise during expert initialization and experimented with different levels of dropout during training. However, the results did not show significant improvements, which led us to develop the expert evolution approach for MoE experts.

\begin{table}[h!]
    \caption{Noise and Dropout Parameters Comparison}
    \label{table:noise_dropout}
    \centering
    \begin{subtable}{0.45\textwidth}
        \centering
        \caption{Noise}
        \label{table:noise}
        \begin{tabular}{l|llll}
            \toprule
            \textbf{Noise} & \textbf{GQA}           & \textbf{SQA}           & \textbf{VQA$^t$}          & \textbf{POPE}          \\ \midrule
            1e-5  & 60.8          & \textbf{63.1} & \textbf{48.0} & \textbf{86.5} \\
            1e-4  & \textbf{61.3} & 62.2          & 47.5          & 86.4          \\
            1e-3  & 61.2          & \textbf{63.1}          & 47.4          & 85.5          \\
            1e-2  & 60.4          & 58.7          & 47.1          & 85.1          \\
            1e-1  & 58.5          & 51.1          & 45.5          & 84.3          \\ \bottomrule
        \end{tabular}
    \end{subtable}
    \hfill
    \begin{subtable}{0.45\textwidth}
        \centering
        \caption{Dropout}
        \label{table:dropout}
        \begin{tabular}{l|llll}
            \toprule
            \textbf{Dropout} & \textbf{GQA}           & \textbf{SQA}           & \textbf{VQA$^t$}          & \textbf{POPE}          \\ \midrule
            0.1     & \textbf{60.6} & 62.1          & 47.5          & 86.1          \\
            0.2     & 60.5          & \textbf{62.2} & 47.4          & 86.2          \\
            0.3     & \textbf{60.6} & 62.1          & \textbf{47.7} & \textbf{86.3} \\
            0.4     & 60.4          & 62.0          & 47.4          & 86.1          \\
            0.5     & 60.3          & 62.0          & 47.2          & 86.2          \\ \bottomrule
        \end{tabular}
    \end{subtable}
\end{table}

\textbf{Exploring the Selection of Evolution Value $\beta$.} To further explore the selection of the evolution value $\beta$, we first fixed $\beta$,and individually evaluated its impact on each expert. As shown in Table 1, when $\beta$ > 0.5, different values perform optimally on their respective benchmarks. In contrast, when $\beta$ < 0.5, most experts exhibit update frequencies similar to the FFN with $\beta$ = 0, resulting in performance comparable to the $\beta$ = 0 case. To achieve stronger generalization capabilities, we opted not to fix $\beta$. Instead, we randomly selected $\beta$ within a specific range at each training step. For instance, as demonstrated in Table 1, randomly choosing $\beta$ between 0.9 and 0.99 yielded the best experimental results. Therefore, in our study, we randomly select $\beta$ from multiple ranges to generate multiple evolved experts.

\begin{table}[!h]
\caption{Evolution value \textbf{$\beta$}. }
\label{table:dropout}
\centering
\begin{tabular}{ccccc|ccc}
\hline
\multicolumn{1}{c|}{\textbf{$\beta$}} & \textbf{VQAv$^{v2}$}         & \textbf{GQA}           & \textbf{SQA}                               & textbf{VQA$^T$}          & \textbf{POPE}          & \textbf{MME}                                 & \textbf{MMB}           \\ \hline
\multicolumn{1}{c|}{0}    & 76.3          & 61.0          & 62.1                              & 48.2          & 86.4          & 1286.7                              & 59.7          \\
\multicolumn{1}{c|}{0.9}  & 76.8          & 60.8          & 62.7                              & 48.6          & \textbf{87.3} & 1290.7                              & 58.4          \\
\multicolumn{1}{c|}{0.8}  & 76.4          & 60.9          & 62.4                              & \textbf{49.0} & 86.6          & 1277.3                              & 61.4          \\
\multicolumn{1}{c|}{0.7}  & \textbf{76.9} & 61.0          & {62.8}                     & 48.7          & 86.4          & 1284.5                              & 59.5          \\ \hline
\multicolumn{1}{c|}{0.4}  & 76.3          & 61.0          & 62.2                              & 48.3          & 86.5          & 1285.9                              & 59.7          \\
\multicolumn{1}{c|}{0.2}  & 76.3          & 61.0          & 62.1                              & 48.2          & 86.4          & 1286.1                              & 59.6          \\ \hline
\multicolumn{1}{c|}{[0.9 - 0.99]} & 76.7          & \textbf{61.1} & \multicolumn{1}{l}{\textbf{63.0}} & 48.9         & 87.0          & \multicolumn{1}{l}{\textbf{1300.7}} & \textbf{61.5} \\ \hline
\end{tabular}
\end{table}



\begin{figure}[t!]
    \centering
    \begin{subfigure}{1\textwidth}
        \centering
        \includegraphics[width=\linewidth]{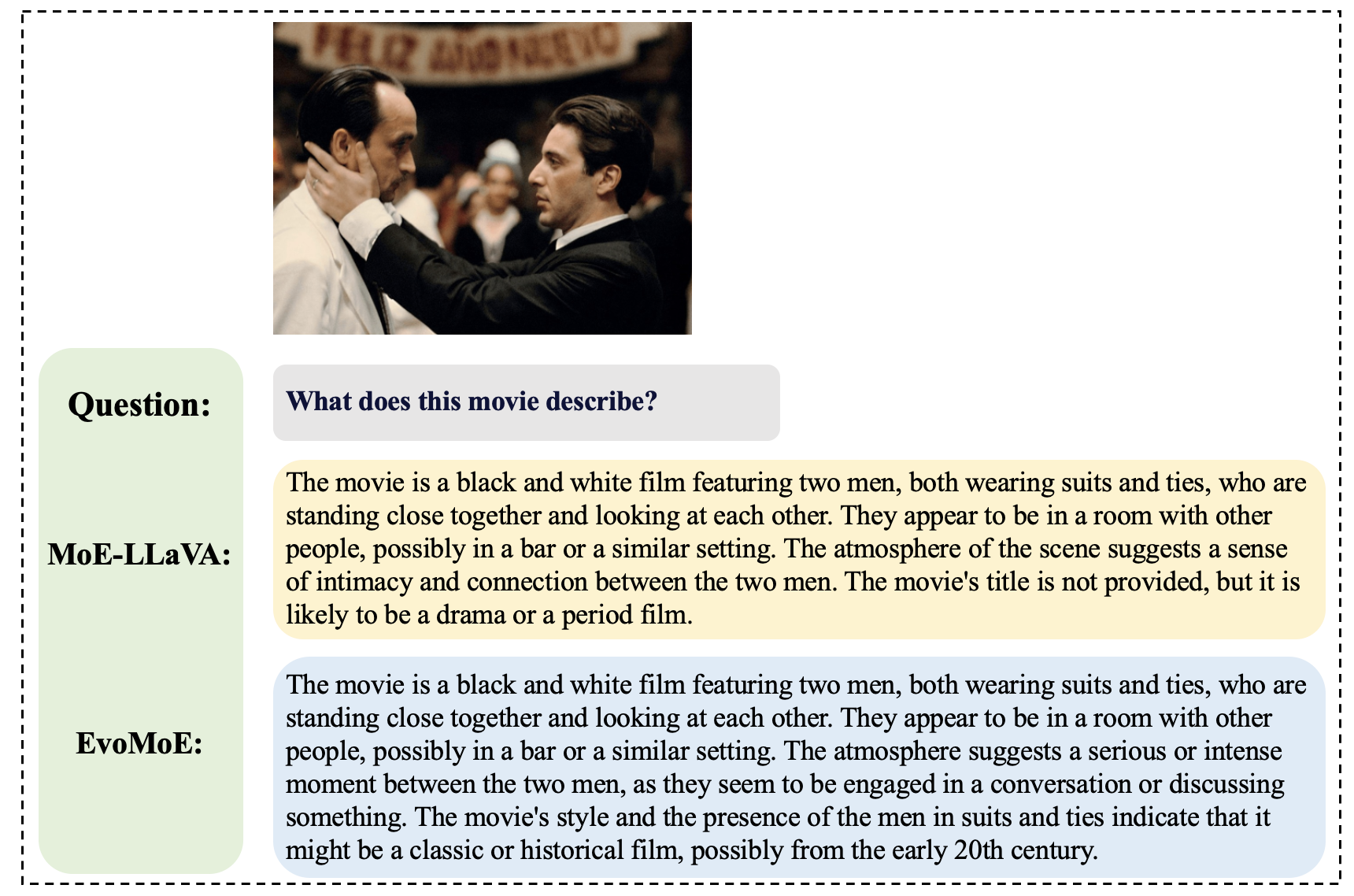} 
    \end{subfigure}
    \hfill
    \begin{subfigure}{1\textwidth}
        \centering
        \includegraphics[width=0.99\linewidth]{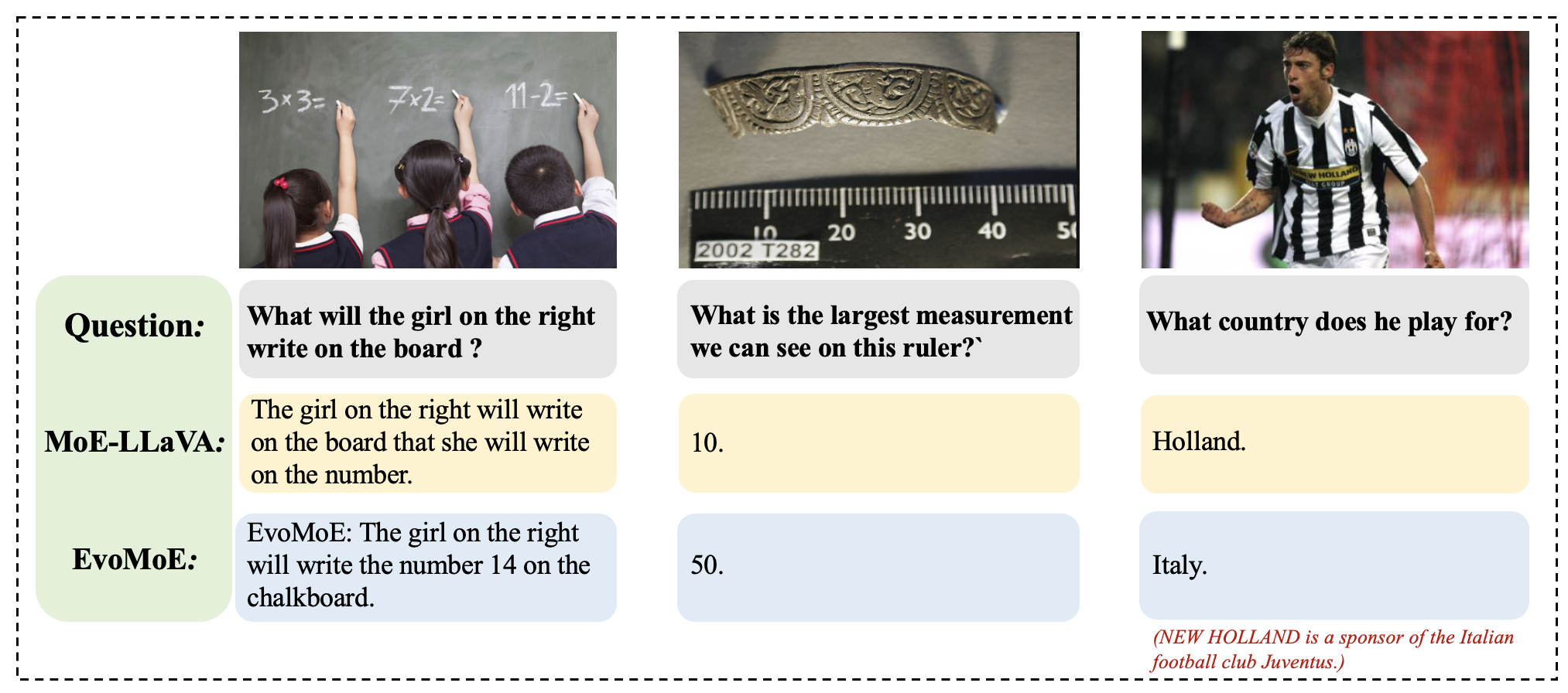} 
    \end{subfigure}
   \caption{Visual input examples.}
    \label{fig:visualizaiton}
\end{figure}

\textbf{Visualization.} In Figure~\ref{fig:visualizaiton}, we present some VQA examples to demonstrate the capabilities of EvoMoE.

\end{document}